\documentclass[lettersize,journal]{IEEEtran}
\usepackage{amsmath,amsfonts}
\usepackage{algorithmic}
\usepackage{algorithm}
\usepackage{array}
\usepackage[caption=false,font=normalsize,labelfont=sf,textfont=sf]{subfig}
\usepackage{textcomp}
\usepackage{stfloats}
\usepackage{url}
\usepackage{verbatim}
\usepackage{graphicx}
\usepackage{cite}
\usepackage{amssymb}
\usepackage{multirow}
\usepackage{array}
\usepackage{graphbox}
\usepackage{booktabs} 
\usepackage{wrapfig}
\usepackage{enumitem}
\usepackage{bbding}
\usepackage{caption}
\usepackage{makecell}
\usepackage{mathrsfs}
\usepackage{color}
\usepackage{colortbl}
\usepackage{xcolor}
\definecolor{mygray}{gray}{.9}
\usepackage[bookmarks=false]{hyperref}
\hypersetup{
    colorlinks = true,
    citecolor  = blue,
    linkcolor  = blue,
    urlcolor   = blue,
}
\usepackage{wrapfig}
\usepackage{enumitem}
\usepackage{listings}
\usepackage{pifont}  % 导入 pifont 宏包
\usepackage{algorithm}
\usepackage{algorithmic}
\usepackage{amsmath}
\newcommand{\tipparagraph}[1]{\vspace{0.2cm}\noindent\textbf{#1}\hspace{0.1cm} }
\usepackage{bm}
\usepackage{booktabs}
\usepackage[table]{xcolor}
\usepackage{multirow}
\usepackage{amssymb}
\definecolor{tabhighlight}{HTML}{e5e5e5}

\newcommand{\tablestyle}[2]{\setlength{\tabcolsep}{#1}\renewcommand{\arraystretch}{#2}\centering\footnotesize}

\begin{document}

\newcommand{\bignum}[1]{\uppercase\expandafter{\romannumeral#1}}

\title{Multi-View Synergistic Learning with Vision-Language Adaption for Low-Resource Biomedical Image Classification}

\author{Xiaoliu Luo, Minxue Xiao, Ting Xie, Mengzhu Wang, \\
Huiqing Qi, Joey Tianyi Zhou, Taiping Zhang, Xu Wang* 

\IEEEcompsocitemizethanks{
\IEEEcompsocthanksitem  X.~Luo, M.~Xiao, and T.~Xie are with Chongqing University of Technology.
\IEEEcompsocthanksitem X.~Wang is with Collge of Computer Science, Sichuan University.
\IEEEcompsocthanksitem M.~Wang is with Hebei University of Technology.
\IEEEcompsocthanksitem H.~Qi is with Nanyang Technological University.
\IEEEcompsocthanksitem X.~Luo, J.~Zhou and X. Wang are with Centre for Frontier AI Research, Agency for Science, Technology and Research, Singapore.
\IEEEcompsocthanksitem T.~Zhang is with Chongqing University.
\IEEEcompsocthanksitem $^*$Corresponding author.

}% <-this % stops an unwanted space
}
        % <-this % stops a space
%\thanks{This paper was produced by the IEEE Publication Technology Group. They are in Piscataway, NJ.}% <-this % stops a space
%\thanks{Manuscript received April 19, 2021; revised August 16, 2021.}}



\maketitle

\begin{abstract}
Accurate biomedical image classification under low-resource conditions remains challenging due to limited annotations, subtle inter-class visual differences, and complex disease semantics. While vision--language models offer a promising foundation for mitigating data scarcity, their effective adaptation in biomedical settings is constrained by the need for parameter-efficient tuning alongside fine-grained and semantically consistent representation learning.
In this work, we propose Multi-View Synergistic Learning (MVSL), a unified framework that addresses these challenges by jointly considering adaptation paradigms, representation granularity, and disease semantic relationships. MVSL decouples the adaptation of visual and textual encoders to respect their distinct representational characteristics, enabling more stable and effective parameter-efficient fine-tuning. It further introduces multi-granularity contrastive learning to explicitly model both global image semantics and localized lesion-level evidence, improving fine-grained discrimination for visually similar disease categories. In addition, MVSL preserves disease-level semantic structure by incorporating structured supervision derived from large language models, which constrains textual representations at the class level and indirectly regularizes visual embeddings through cross-modal alignment.
Together, these components enable more stable cross-modal alignment and improved discrimination under limited supervision.
Extensive experiments on $11$ public biomedical datasets spanning $9$ imaging modalities and $10$ anatomical regions demonstrate that MVSL consistently outperforms state-of-the-art methods in few-shot and zero-shot classification settings. Our code will be available at  \href{https://github.com/luoxiaoliu/MVSL}{Github}. 
\end{abstract}

%%%%%%%%% BODY TEXT
\section{Introduction}
\label{sec:intro}
Biomedical image classification under low-resource conditions remains a persistent challenge in medical image analysis. In clinical practice, annotated data are often scarce due to the rarity of certain diseases, the cost of data acquisition, and the reliance on expert-level annotation. These difficulties are further compounded by the subtle visual differences and complex semantic relationships that characterize many biomedical categories, making reliable recognition particularly challenging in few-shot or zero-shot settings.

%%%%%%%%%%%%%%%%%%%%%
\begin{figure}[!t]
\centering
    \begin{minipage}   {0.99\linewidth}
        \centering
        \includegraphics [width=1\linewidth] 
        {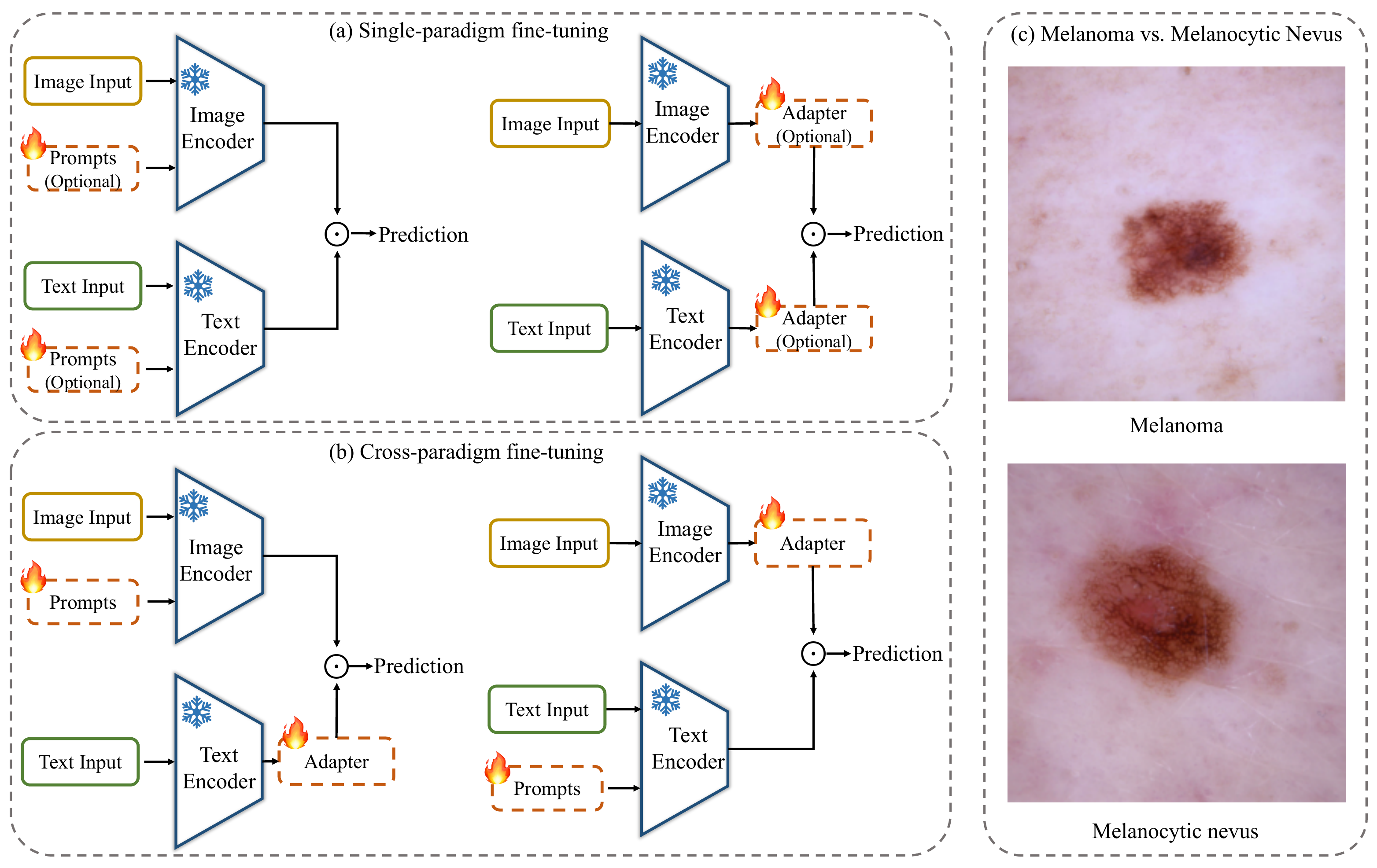}
    \end{minipage}
    %\vspace{-0.2cm}
    \caption{Illustration of single- and cross-paradigm fine-tuning strategies for vision-language models, alongside examples of visually similar skin lesions. (a) Single-paradigm fine-tuning applies the same adaptation method (prompt tuning or adapter tuning) to both image and text encoders. (b) Cross-paradigm fine-tuning uses different adaptation methods for each modality, enabling more flexible representation learning. (c) Example images of melanoma and melanocytic nevus, highlighting the challenge of distinguishing visually similar disease categories.}
   %\vspace{-0.2cm}
    \label{fig:fine-tuning}
\end{figure}
%%%%%%%%%%%%%%%%%%%%%
Recent advances in vision--language models (VLMs) have shown strong potential for alleviating data scarcity by leveraging large-scale cross-modal pretraining. Models such as CLIP~\cite{CLIP}, ALIGN~\cite{ALIGN}, LiT~\cite{zhai2022lit}, and FILIP~\cite{FILIP} learn unified embedding spaces through contrastive image--text supervision, enabling impressive zero-shot and open-set recognition performance in natural image domains. Building on these successes, growing efforts have focused on adapting VLMs to the medical domain in order to exploit abundant unstructured clinical text alongside medical images. Biomedical VLMs, including BiomedCLIP~\cite{zhang2025biomedclip}, PubMedCLIP~\cite{pubmedclip}, and PMC-CLIP~\cite{PMC-CLIP}, are pretrained on large-scale medical image--text pairs and capture rich associations among diseases, imaging modalities, and clinical terminology, providing a strong foundation for downstream medical image analysis.

Despite these advances, effectively adapting pretrained biomedical VLMs to low-resource classification tasks remains an open problem. Most existing approaches rely on parameter-efficient fine-tuning strategies, such as CoOp~\cite{CoOp}, CLIP-Adapter~\cite{clip_adapter}, and LP++~\cite{huang2024lp++}, which introduce a small number of trainable parameters to mitigate overfitting and reduce computational cost. However, these methods typically adopt a single fine-tuning paradigm that is applied uniformly to both the visual and textual encoders, as illustrated in Figure~\ref{fig:fine-tuning}(a). While appealing in its simplicity, this uniform treatment overlooks important differences between visual and textual representations. Visual encoders process high-dimensional, continuous, and spatially structured features, where fine-grained textures, lesion regions, and spatial relationships play a critical role in medical discrimination. In contrast, text encoders operate on discrete token sequences with abstract and compositional semantics, which are often sensitive to contextual conditioning. Applying the same adaptation strategy to both modalities can therefore lead to suboptimal representation learning and unstable cross-modal alignment, particularly when training data are limited. 

To address this issue, we propose \emph{Cross-Paradigm Fine-Tuning} (CPFT), a modality-aware adaptation strategy that decouples the fine-tuning mechanisms of the visual and textual branches. As shown in Figure~\ref{fig:fine-tuning}(b), CPFT introduces lightweight residual adapters into intermediate layers of the visual encoder to refine spatial and lesion-level representations, while employing learnable prompts to inject task-specific semantic context into the text encoder. This asymmetric design allows each modality to be adapted in a manner consistent with its representational structure, improving intra-modal adaptation while preserving cross-modal alignment.

Although CPFT improves modality-aware adaptation, it does not fully resolve another key limitation of existing VLM-based approaches: the weak correspondence between localized visual evidence and disease semantics. Most contrastive learning methods emphasize global image--text alignment, which can be insufficient when discriminative cues are spatially localized, as is common in biomedical images. To better capture such fine-grained information, we introduce \emph{Multi-Granularity Contrastive Learning} (MGCL), which jointly models global and local correspondences between images and text. MGCL aligns holistic image representations with class-level textual embeddings while simultaneously enforcing patch-level alignment to highlight lesion regions and clinically relevant visual cues. A dynamic fusion mechanism further integrates global and local predictions, resulting in more robust inference under limited supervision.

Even with modality-aware adaptation and multi-granularity alignment, generalization remains challenging for visually similar disease categories. As illustrated in Figure~\ref{fig:fine-tuning}(c), melanoma and melanocytic nevus in DermaMNIST~\cite{dermamnist1,dermamnist2} exhibit highly similar visual appearances, making them difficult to distinguish based on visual features alone. Existing methods typically constrain the textual branch through per-sample alignment objectives, such as minimizing the mean squared error between student and teacher embeddings. While effective at enforcing local consistency, such strategies fail to preserve class-level semantic structure, which is particularly important in low-resource settings and can otherwise lead to scattered intra-class representations or entangled inter-class embeddings. To explicitly model disease-level semantic relationships, we propose the \emph{Disease Semantic Graph} (DSG). DSG leverages large language models to generate class-level textual embeddings and constructs a semantic graph that captures the global topology among disease categories. This graph serves as a teacher prior to guide textual fine-tuning through structural distillation, encouraging semantic consistency within classes and improved separation between visually similar diseases. Through its interaction with CPFT and MGCL, this class-level semantic structure is further propagated to the visual branch, indirectly regularizing visual representations.

In summary, we present \emph{Multi-View Synergistic Learning} (MVSL), a unified framework for low-resource biomedical image classification that integrates modality-aware fine-tuning, multi-granularity representation alignment, and class-level semantic structure modeling. By jointly addressing these complementary aspects, MVSL improves adaptability, discriminability, and generalization under limited supervision. Extensive experiments on 11 public biomedical datasets spanning 9 imaging modalities and 10 anatomical regions demonstrate that MVSL consistently outperforms state-of-the-art methods.

Our main contributions are summarized as follows:
\begin{itemize}
\item We propose a modality-aware cross-paradigm fine-tuning strategy that applies structured adapters to visual encoders and learnable prompts to text encoders, enabling more effective parameter-efficient adaptation of biomedical VLMs.
\item We introduce a multi-granularity contrastive learning framework that jointly aligns global image semantics and localized lesion-level features with textual representations, improving fine-grained discrimination in low-resource settings.
\item We propose a disease semantic graph that captures class-level semantic topology and constrains textual fine-tuning via structural distillation, indirectly regularizing visual representations and enhancing inter-class discriminability.
\item We integrate these components into the MVSL framework and demonstrate its effectiveness through extensive experiments on diverse biomedical benchmarks, achieving consistent improvements over state-of-the-art approaches.
\end{itemize}

\section{Related Work}

\subsection{VLMs in General and Biomedical Domains}
Vision-Language Models (VLMs) are a class of cross-modal representation learning models capable of jointly understanding images and text, with core components including an image encoder, a text encoder, and corresponding training objectives. Foundational models such as CLIP \cite{CLIP}, ALIGN \cite{ALIGN}, LiT \cite{zhai2022lit}, and FILIP \cite{FILIP} are pretrained on large-scale image-text pairs to map images and text into a shared vector space, achieving semantic alignment. This enables the models to encode high-dimensional image and text information into comparable feature vectors and perform zero-shot prediction and cross-modal tasks via similarity measurement. In the natural image domain, VLMs have been widely applied to tasks such as image captioning, cross-modal retrieval~\cite{peng2026semantic,Wang2023CoDA,10119165,su2025dica,su2025neighbor,yin2025roda} , and visual question answering, enhancing both multimodal understanding and interaction efficiency. Even when certain domain-specific concepts are absent from pretraining data, models can still reason using related semantic information.

However, general VLMs face performance limitations in specialized domains such as biomedical imaging, primarily due to the substantial semantic gap between pretraining data and target domains. To address this challenge, researchers have developed domain-specific biomedical models. For instance, PMC-CLIP \cite{PMC-CLIP} provides an automated image-text data construction pipeline and the PMC-OA dataset, though the model architecture is not specifically optimized. PubMedCLIP \cite{pubmedclip} applies domain-adaptive fine-tuning on PubMed literature, adjusting only the loss function to better capture biomedical semantics. BiomedCLIP \cite{zhang2025biomedclip} further improves both data scale and model architecture, with a training set far exceeding existing biomedical multimodal datasets, a BERT-based text encoder with extended context window, and a larger ViT image encoder to process high-resolution images, enabling more precise modeling of complex biomedical semantic relationships.

Although these models perform well on standard benchmarks, they still require task-specific fine-tuning to handle the fine-grained nature of biomedical image analysis, particularly in low-resource settings. This challenge has motivated the exploration of parameter-efficient fine-tuning strategies that support rapid domain adaptation, strengthen semantic alignment, and improve few-shot generalization in downstream medical applications.

\subsection{Low-Resource Biomedical Image Classification}
Learning under low-resource conditions has become a fundamental challenge in computer vision, where limited annotations, distribution shifts, and semantic ambiguity hinder reliable model generalization. To address these issues, various learning paradigms have been extensively studied from different perspectives. Few-shot learning~\cite{tian2020pfenet,10148619,10672530,9921265,10305430,hsrap,DAI2023109108,10551867,LUO2023108811} enhances intra-class compactness and inter-class separability to enable rapid adaptation with limited labeled samples. Semi-supervised learning~\cite{Wang2020DeepSC} leverages both labeled and unlabeled data, exploiting consistency constraints and intrinsic data structures to improve representation learning. Domain generalization methods further aim to learn representations that remain robust to unseen domain shifts, and recent prompt- and contrastive-learning based approaches have shown that misleading domain cues and pseudo-label contrastive objectives can improve domain-invariant representation learning~\cite{li2026GenPromptCL}. Zero-shot learning relies on semantic priors and cross-modal alignment to transfer knowledge from seen to unseen categories. Collectively, these approaches alleviate the low-resource bottleneck from the perspectives of data utilization, structural exploitation, and semantic transfer. 

When applied to biomedical image classification, however, these methods face additional challenges due to the unique characteristics of medical data. Compared with natural images, biomedical images typically exhibit small and sparsely distributed lesions, substantial intra-class variability, and subtle inter-class differences. Moreover, disease categories often involve complex and implicit semantic relationships that are difficult to capture with conventional representation learning methods. 

Recently, VLM-based approaches have demonstrated substantial advantages in low-resource biomedical image classification by incorporating cross-modal knowledge and semantic guidance from textual descriptions. Prompt learning has emerged as an effective and parameter-efficient strategy. Methods such as CoOp~\cite{CoOp} and CoCoOp~\cite{cocoop} employ learnable prompts to condition text encoders for specific tasks, while advanced approaches like ProDA~\cite{ProDA}, ProGrad~\cite{ProGrad}, Bayesian Prompt Learning~\cite{bayesian}, and PromptSRC~\cite{PromptSRC}, CLSEP~\cite{WANG2023110381} enhance prompt robustness and semantic grounding. In the biomedical domain, approaches including ViP~\cite{fang2024aligning}, XCoOp~\cite{bie2024xcoop}, and DCPL~\cite{cao2024domain} further integrate medical terminology and domain-specific biases. LLM-assisted methods, such as ProText~\cite{Khattak2024ProText} and BiomedCoOp~\cite{koleilat2025biomedcoop}, provide automated prompt generation and refinement, though their integration with biomedical-specific VLMs remains limited.

Adapter-based tuning provides a complementary pathway by introducing lightweight modules into frozen VLM backbones. CLIP-Adapter~\cite{clip_adapter} inserts trainable layers into the visual encoder, Tip-Adapter~\cite{zhang2021tip} combines support-set features with pretrained embeddings, and LP++~\cite{huang2024lp++} and CLAP~\cite{silva2024closer} regulate adaptation via implicit learning rates or regularization toward pretrained prototypes. Adapter-based approaches allow structural intervention in the visual pipeline, particularly beneficial for domain-specific visual patterns. However, these methods often lack semantic conditioning, and applying them uniformly across modalities may compromise cross-modal generalization.

Despite these advances, existing VLM-based approaches often optimize only text-level or global visual features, leaving fine-grained alignment between localized lesions and textual descriptions underexplored. Modeling inter-disease semantic relationships and local cross-modal correspondences remains limited, especially for rare or visually similar disease categories. Motivated by these limitations, we propose Multi-View Synergistic Learning (MVSL), which integrates heterogeneous fine-tuning strategies, multi-granularity feature alignment, and disease semantic relationship modeling to address cross-modal misalignment and enhance fine-grained lesion representation in low-resource biomedical image classification, achieving more accurate and robust performance.

%%%%%%%%%%%%%%%%%%%%%
\begin{figure*}[!t]
\centering
    \begin{minipage}   {0.99\linewidth}
        \centering
        \includegraphics [width=1\linewidth] 
        {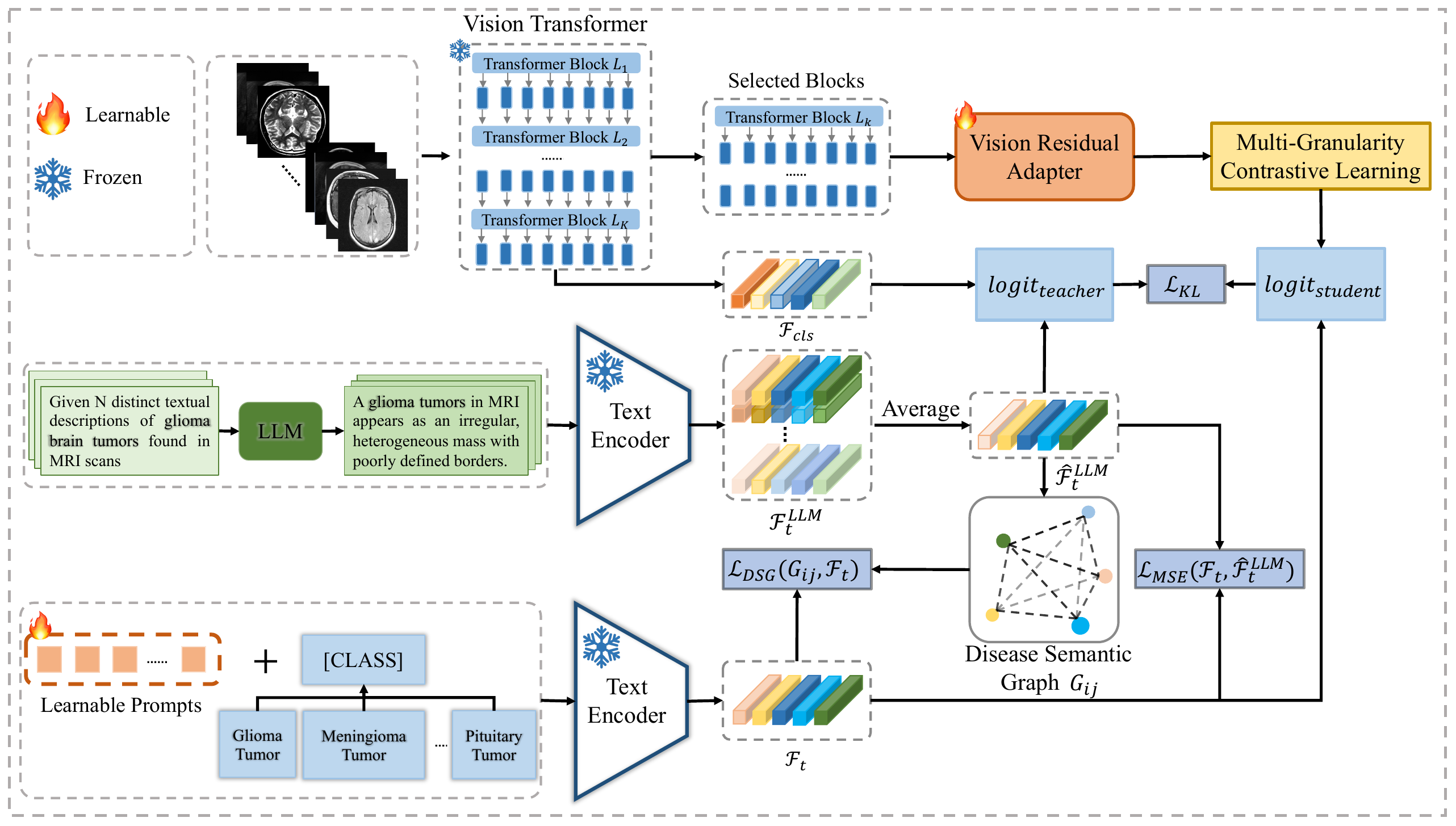}
    \end{minipage}
    \vspace{-0.2cm}
    \caption{Overview of the proposed MVSL framework. MVSL unifies fine-tuning paradigms, representation granularity, and disease semantic relationships into a cohesive framework. In cross-paradigm fine-tuning, the adaptations of the visual and text encoders are decoupled to respect their distinct representational characteristics, enabling more stable and parameter-efficient optimization. Class-specific medical descriptions generated by LLM are leveraged to construct a disease semantic graph and, via distillation, constrain the prompt text representations to preserve disease-level semantic structure, while simultaneously regularizing visual embeddings through cross-modal alignment. Multi-granularity contrastive learning explicitly captures both global image semantics and local lesion-level evidence, enhancing fine-grained discrimination among visually similar disease categories. Finally, predictions obtained across multiple granularities are dynamically fused to deliver robust few-shot and zero-shot biomedical image classification performance.}
    \vspace{-0.2cm}
    \label{fig:arch}
\end{figure*}
%%%%%%%%%%%%%%%%%%%%%

\section{Method}
In this section, we introduce the Multi-View Synergistic Learning (MVSL) framework, a unified approach designed for low-resource biomedical image classification. MVSL systematically integrates three complementary perspectives: fine-tuning paradigms, visual representations, and disease semantic relationships, to enhance the adaptation of pretrained biomedical VLMs, such as BiomedCLIP, under few-shot and zero-shot settings. Specifically, Cross-Paradigm Fine-Tuning (CPFT) independently optimizes the visual and textual branches, enabling fine-grained modulation of local lesion features and semantic information. Multi-Granularity Contrastive Learning (MGCL) explicitly models global and local correspondences between images and text, improving classification performance for scarce classes. The Disease Semantic Graph (DSG) captures class-level semantic topology, constraining the textual embeddings and indirectly regularizing visual embeddings to maintain discriminability among visually similar disease categories. The synergistic integration of these components allows MVSL to fully leverage pretrained models for robust downstream medical image classification. An overview of the MVSL architecture is shown in Figure~\ref{fig:arch}, with each module and its function described in the following subsections.

\subsection{Cross-Paradigm Fine-Tuning}
To efficiently adapt pretrained biomedical vision–language models to downstream classification tasks under low-resource settings, we propose the Cross-Paradigm Fine-Tuning (CPFT) strategy, which explicitly leverages the complementary characteristics of visual and textual modalities to perform modality-specific adaptations while preserving pretained knowledge. 

The visual encoder processes high-dimensional, continuous spatial features with locally dense and hierarchical structures, making it well-suited for fine-grained modulation via structured adapters. In contrast, the text encoder operates over discrete symbol sequences with highly abstract, context-dependent semantics, making it more suitable for semantic conditioning through learnable prompts. Based on this insight, as illustrated in Figure~\ref{fig:arch}, CPFT exploits these difference by adopting an asymmetric, modality-aware fine-tuning schemes: structured adapters for the visual branch and prompt-based conditioning for the textual branch. This design enables both branches to adapt efficiently without disrupting the pretained cross-modal alignment.

\begin{figure}[!t]
\centering
    \begin{minipage}   {0.99\linewidth}
        \centering
        \includegraphics [width=1\linewidth] 
        {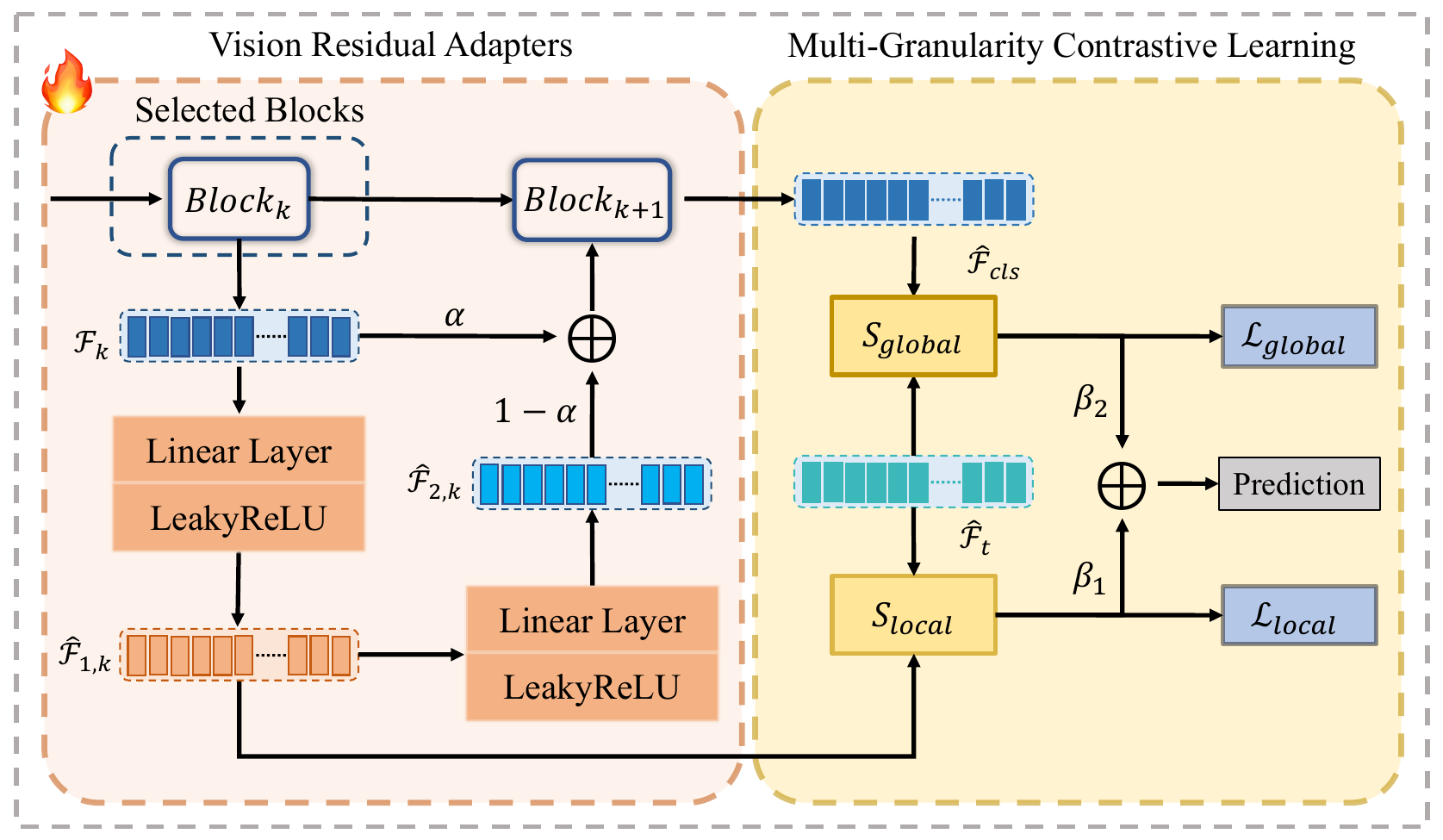}
    \end{minipage}
    \vspace{-0.2cm}
    \caption{Detailed architecture of the VRA and MGCL modules. The Visual Residual Adapter (VRA) applies lightweight residual adapter to selected blocks of the frozen visual encoder, enabling fine-grained adaptation of visual features. The Multi-Granularity Contrastive Learning (MGCL) module aligns both global and local correspondences between the visual and textual branches, and a dynamic fusion mechanism integrates these granularities to enhance cross-modal alignment and boost few-shot and zero-shot classification performance.}
    \vspace{-0.2cm}
    \label{fig:vra}
\end{figure}

\tipparagraph{Vision Residual Adapter.} Building on the visual adaptation strategy of CPFT, we introduce the Vision Residual Adapter (VRA) to achieve fine-grained modulation of visual features. Biomedical images involve subtle morphological cues and high intra-class variability, which necessitates careful structural refinement. To accomplish this without compromising the stability of the pretrained backbone, lightweight residual adapters are inserted into selected Transformer blocks of the frozen ViT model. As shown in Figure~\ref{fig:vra}, each adapter follows a bottleneck design: visual features are projected into a lower-dimensional latent space, transformed through a nonlinear activation, and then projected back to the original feature dimension. This enables the adapter to learn fine-grained, task-specific corrections to the visual representation.

Formally, let $F_{\text{cls}} \in \mathbb{R}^{B \times D}$ denote the final outputs of the frozen vision encoder $E_v(\cdot)$, where $B$ is the batch size, $D$ is the final embedding dimension. Let $\mathcal{F}_k \in \mathbb{R}^{B \times N \times d}$ denote the output of the $k$-th selected transformer block from the frozen vision encoder, where $N$ is the number of visual tokens, and $d$ is the bottleneck embedding dimension. The adapter performs a two-step transformation:
\begin{equation}
\hat{\mathcal{F}}_{1,k} = \text{ReLU}(\mathcal{F}_k W_{1,k}),
\end{equation}
\begin{equation}
\hat{\mathcal{F}}_{2,k} = \text{ReLU}(\hat{\mathcal{F}}_{1,k} W_{2,k}),
\end{equation}
where $W_{1,k} \in \mathbb{R}^{d \times D}$ and $W_{2,k} \in \mathbb{R}^{D \times d}$ are trainable projection matrices with $D < d$. 

The adapted feature is then combined with the original backbone output through a scaled residual connection:
\begin{equation}
\hat{\mathcal{F}}_k = \alpha \mathcal{F}_k + (1 - \alpha) \hat{\mathcal{F}}_{2,k},
\end{equation}
where $\alpha$ controls the contribution of the adapter. Only the adapter parameters are trainable, the backbone remains frozen. This design provides a parameter-efficient yet expressive way to adjust the visual encoder to biomedical classification needs while maintaining the spatial structure and robustness learned during pre-training. Finally, to distinguish the adapted features form the frozen ones, we denote the final visual feature output after applying adapters as $\hat{\mathcal{F}}_{\text{cls}} \in \mathbb{R}^{B \times D}$.

\tipparagraph{Text Prompt Tuning.} While the visual branch relies on structural adaptation, the textual branch requires a complementary approach due to its discrete and highly abstract semantic nature. Biomedical texts, including disease names, phenotypic descriptions, and anatomical terminology, contain dense semantic information but have relatively simple structural patterns. Therefore, instead of modifying the pretrained text encoder $E_t$, CPFT introduces learnable prompts to inject task- or domain-specific knowledge. 
%%%%%%%%%%%%%%%%%%%%%
%%%%%%%%%%%%%%%%%%%%%

Formally, let $F_{t} \in \mathbb{R}^{C \times D}$ denote the final outputs of the frozen text encoder, where $C$ is the number of disease class. Following the CoOp paradigm, for each disease class $c$, let $T_c=\{t_{1,c},\dots,t_{L,c}\}$ denote the label token sequence, where $L$ is the length of the label description. We prepend $M$ learnable context tokens $P=\{p_1, \dots, p_m\}$ to form the prompted input:
\begin{equation}
\hat{T}_c = \{p_1,\dots,p_M, t_{1,c}, \dots, t_{L,c}\}.
\end{equation}
These prompts are optimized jointly with the task objectives while keeping the text encoder frozen, serving as soft domain descriptors to enrich biomedical semantic cues, such as lesion attributes, modality characteristics, or anatomical priors, into the encoder’s embedding space. The resulting text embedding $\hat{f}_{t,c} \in \mathbb{R}^{1 \times D}$ is:
\begin{equation}
\hat{f}_{t,c} = E_{t}(\hat{T}_c).
\end{equation}
To distinguish the tuned features form the frozen ones, we denote the final prompt-tuned text features as $\hat{\mathcal{F}}_{t} = \{\hat{f}_{t,c}\}_{c=1}^{C} \in \mathbb{R}^{C \times D}$.

To further enhance semantic richness and ensure alignment with biomedical terminology, MVSL follows the BiomedCoOp paradigm by leveraging large language models (LLMs) to generate diverse prompts. To overcome the limitations of manual prompt engineering in capturing nuanced visual cues, we query the LLM with the instruction: \textit{“Give N textual descriptions of visual discriminative features for distinct medical cases of [CLASS] found in [MODALITY].”}. This formulation guides the LLM to produce prompts that are both clinically accurate and modality-consistent. For the $i$-th disease class, the generated prompt set is denoted as $x_i \in \mathbb{R}^{\text{Num} \times D}$, where $\text{Num}$ indicates the number of prompts. These are subsequently encoded by the text encoder to yield fixed text embeddings $F_t^{LLM} =E_t{(\{x_i\}_{i=1}^C)} \in \mathbb{R}^{\text{Num} \times C \times D}$, providing additional supervision to guide downstream classification.
%%%%%%%%%%%%%%%%%%%%%%%%

\subsection{Multi-Granularity Contrastive Learning}
Although CPFT enhances modality-aware adaptation, it does not fully address a fundamental limitation of existing VLMs-based approaches for low-resource biomedical image classification, namely the weak alignment between localized visual evidence and disease semantics. In biomedical images, discriminative cues are often subtle and spatially localized, such as lesion boundaries or pathological regions, while textual descriptions typically encode rich entity-level semantic information, including anatomical locations, disease types, and pathological attributes. However, conventional contrastive learning frameworks mainly emphasize global image–text alignment, which may overlook fine-grained correspondences between local visual patterns and clinically relevant semantic entities, thereby limiting discrimination among disease categories.

To address this issue, we propose Multi-Granularity Contrastive Learning (MGCL), which explicitly models image–text correspondences at both global and local granularities. MGCL jointly aligns holistic visual representations with class-level textual embeddings while simultaneously enforcing localized alignment between image patches and disease semantics. This multi-granularity formulation enables the model to capture fine-grained visual evidence and its semantic associations, improving robustness and generalization under limited supervision.

At the global level, MGCL enforces semantic consistency between modalities by aligning adapted visual representations with prompt-tuned textual embeddings. Specifically, the final visual features after adapter tuning,  denoted as $\hat{F}_{cls} \in \mathbb{R}^{B \times D}$, are aligned with the corresponding class-level textual representations $\hat{F}_t \in \mathbb{R}^{C \times D}$ in a shared embedding space. This alignment encourages each image to be strongly associated with its disease category at a semantic level. We first compute the cosine similarity matrix between image and text embeddings:

\begin{equation}
S_{\text{global}} = \frac{\hat{F}_{\text{cls}} \hat{F}_t^\top}{\|\hat{F}_{\text{cls}}\| \, \|\hat{F}_t\|} \in \mathbb{R}^{B \times C},
\end{equation}
where $\hat{F}_t^\top$ denotes the transpose of $\hat{F}_t$, and the norms are applied row-wise for normalization. Based on this similarity matrix, the global contrastive loss is defined as:

\begin{equation}
\mathcal{L}_{\text{global}} = - \frac{1}{B} \sum_{i=1}^{B} \log \frac{\exp(S_{\text{global}}[i, y_i]/\tau)}{\sum_{j=1}^{C} \exp(S_{\text{global}}[i,j]/\tau)},
\end{equation}
where $y_i$ is the ground-truth class index of the $i$-th image, $\tau$ is a temperature parameter, and $S_{\text{global}}[i,j]$ is the cosine similarity between image $i$ and class $j$. This objective promotes high similarity between each image and its correct class embedding while suppressing similarities to other classes, thereby capturing high-level disease semantics and providing a stable foundation for cross-modal representation learning.

Beyond global alignment, MGCL further incorporates local-level contrastive learning to highlight lesion regions and clinically informative visual cues. By enforcing patch-level alignment between localized visual features and textual semantics, the model is encouraged to focus on discriminative regions that are critical for differentiating visually similar disease categories. Concretely, we extract patch-level features $\hat{\mathcal{F}}_{1,k,p} \in \mathbb{R}^{B \times (N-1) \times D}$ from the first-step adapter-transformed representations $\hat{\mathcal{F}}_{1,k}$. These patch features are then aligned with class-level textual embeddings $\hat{F}_t \in \mathbb{R}^{C \times D}$. The patch-text similarity matrix is computed as:
\begin{equation}
S_{\text{patch}} = \frac{\hat{\mathcal{F}}_{1,k,p} \hat{F}_t^\top}{\|\hat{\mathcal{F}}_{1,k,p}\| \, \|\hat{F}_t\|} \in \mathbb{R}^{B \times (N-1) \times C},
\end{equation}
where normalization is applied at the patch level. The corresponding local contrastive loss is formulated as:
\begin{equation}
\mathcal{L}_{\text{local}} = - \frac{1}{B(N-1)} \sum_{i=1}^{B} \sum_{n=1}^{N-1} \log \frac{\exp(S_{\text{patch}}[i,n, y_i]/\tau)}{\sum_{c=1}^{C} \exp(S_{\text{patch}}[i,n,c]/\tau)},
\end{equation}
where $S_{\text{patch}}[i,n,c]$ denotes the cosine similarity between the $n$-th patch of image $i$ and class $c$, $\tau$ is a temperature parameter, and $y_i$ is the ground-truth class. This formulation enforces each local region to align with its corresponding disease semantics, thereby strengthening fine-grained cross-modal correspondence.

Finally, MGCL integrates predictions derived from different granularity levels to produce robust inference. Patch-level similarities are first aggregated across spatial locations to obtain a single local prediction per image:
\begin{equation}
S_{\text{local}} = \frac{1}{N-1} \sum_{n=1}^{N-1} S_{\text{patch}}[:,n,:] \in \mathbb{R}^{B \times C}.
\end{equation}

The final prediction is computed as a weighted combination of global and local predictions:

\begin{equation}
S_{\text{final}} = \beta_1 \cdot S_{\text{local}} + \beta_2 \cdot S_{\text{global}},
\end{equation}
where $\beta_1$ and $\beta_2$ are learnable fusion coefficients that control the contribution of local and global information, respectively. This dynamic fusion mechanism enables the model to adaptively combine holistic contextual understanding with fine-grained lesion evidence, leading to more robust and discriminative predictions, particularly under low-resource conditions or for diseases with subtle or localized manifestations. 

\subsection{Disease Semantic Graph}
Despite CPFT and MGCL facilitating intra-modal adaptation and local cross-modal alignment, models still face two major challenges in low-resource settings. First, visually similar disease categories, such as melanoma and melanocytic nevus in DermaMNIST, are difficult for the visual branch to distinguish due to subtle differences in local texture patterns. Second, existing approaches to constrain the textual branch typically rely on per-sample alignment objective, such as minimizing the mean squared error between student and teacher textual embeddings, which enforce local consistency but ignore class-level semantic structures. Neglecting these structures can lead to scattered intra-class embeddings or overlapping inter-class embeddings, thereby limiting cross-modal discriminability and overall generalization. 

To address this, we propose the Disease Semantic Graph (DSG), which explicitly models class-level semantic relationships and guides the fine-tuning of the textual branch via structural distillation. Specifically, prompts generated by LLMs are encoded by the text branch to produce class-level textual embeddings. Let there be $C$ disease categories, and denote the embeddings generated from each class’s prompts as $F_t^{\text{LLM}} \in \mathbb{R}^{\text{Num} \times C \times D}$, where $\text{Num}$ is the number of prompts per class. The class-level text embedding is obtained by averaging across the prompt dimension:
\begin{equation}
\hat{F}_t^{\text{LLM}} = \frac{1}{\text{Num}} \sum_{i=1}^{\text{Num}} F_t^{\text{LLM}}[i, :, :].
\end{equation}

Next, the semantic relationships among disease categories are encoded in a soft adjacency matrix $G \in \mathbb{R}^{C \times C}$, computed based on pairwise cosine similarity:
\begin{equation}
G_{ij} = \frac{\exp \left( \text{cos} \big( \hat{F}_t^{\text{LLM}}[i,:], \hat{F}_t^{\text{LLM}}[j,:] \big) / \tau \right)}
{\sum_{c=1}^{C} \exp \left( \text{cos} \big( \hat{F}_t^{\text{LLM}}[i,:], \hat{F}_t^{\text{LLM}}[c,:] \big) / \tau \right)},
\end{equation}
where $\text{cos}(\cdot,\cdot)$ denotes cosine similarity, and $\tau$ is a temperature parameter controlling the sharpness of the distribution. This matrix captures the global semantic topology among disease classes, serving as a teacher prior for the student model.

The student textual embeddings $\hat{F}_t$ are then regularized via Laplacian structural distillation:
\begin{equation}
\mathcal{L}_{\text{DSG}} = \frac{1}{C^2} \sum_{i=1}^{C} \sum_{j=1}^{C} G_{ij} \cdot \| \hat{F}_t[i,:] - \hat{F}_t[j,:] \|_2^2.
\end{equation}

Minimizing $\mathcal{L}_{\text{DSG}} $ encourages the student model to preserve the teacher embeddings’ semantic topology, maintaining intra-class consistency and inter-class discriminability. Furthermore, since the visual branch is aligned with textual embeddings through CPFT and MGCL, this class-level semantic structure indirectly regularizes visual representations, improving discriminability among visually similar diseases and boosting generalization to rare or low-resource classes.

\subsection{Objective Function}
The overall training objective is designed to jointly optimize multi-granularity visual–text contrastive learning, structural guidance from the DSG, and teacher-student supervision derived from pretrained and prompt-enhanced embeddings. Specifically, the frozen vision encoder serves as the vision teacher, producing image-level representations denoted as $F_{\text{cls}}$. Similarly, the frozen text encoder serves as the text teacher, with teacher logits computed by measuring the cosine similarity between $F_{\text{cls}}$ and the class-level text embeddings $\hat{F}_t^{\text{LLM}}$:

\begin{equation}
\text{logit}_{\text{teacher}} = \text{sim}(F_{\text{cls}}, \hat{F}_t^{\text{LLM}}),
\end{equation}
where $\text{sim}(\cdot, \cdot)$ denotes cosine similarity. 

In parallel, the student model generates adapted image features $\hat{F}_{\text{cls}}$ and prompt-tuned text features $\hat{F}_t$ through CPFT, forming the student logits $\text{logit}_{\text{student}} = \text{sim}(\hat{F}_{\text{cls}}, \hat{F}_t)$. 

Teacher-student alignment is enforced by minimizing the Kullback–Leibler (KL) divergence between their predicted logits:
\begin{equation}
\mathcal{L}_{\text{KL}} = \text{KL} \big( \text{softmax}(\text{logit}_{\text{teacher}} / \tau) \, \| \, \text{softmax}(\text{logit}_{\text{student}} / \tau) \big),
\end{equation}
where $\tau$ is a temperature parameter. Additionally, to ensure semantic consistency between the student text features and the LLM-enhanced text features, a mean squared error (MSE) loss is applied:
\begin{equation}
\mathcal{L}_{\text{MSE}} = \frac{1}{C} \sum_{c=1}^{C} \| \hat{F}_t[c,:] - \hat{F}_t^{\text{LLM}}[c,:] \|_2^2.
\end{equation}
This dual supervision encourages the learned visual–text representations to remain aligned with the pretrained cross-modal space while incorporating high-quality biomedical semantics from LLM-generated prompts.

Finally, the total loss integrates multi-granularity contrastive learning, structural guidance, and teacher-student supervision:

\begin{equation}
\mathcal{L}_{\text{total}} =  \mathcal{L}_{\text{global}} + \mathcal{L}_{\text{local}} + \lambda_1 \mathcal{L}_{\text{MSE}} + \lambda_2 \mathcal{L}_{\text{KL}}  + \lambda_3 \mathcal{L}_{\text{DSG}} ,
\end{equation}
where $\lambda_1,\lambda_2$, and $\lambda_3$ are weighting coefficients controlling the contribution of each component. This unified objective encourages the student model to capture coarse- and fine-grained cross-modal correspondences, preserve disease semantic structures, and mimic the knowledge of teacher embeddings, thereby improving generalization under low-resource biomedical classification scenarios.

\section{Experiments}
We conduct a comprehensive evaluation of the proposed MVSL framework on a diverse set of publicly available biomedical imaging datasets, covering multiple imaging modalities and anatomical regions. Specifically, we evaluate the model in few-shot and zero-shot settings to simulate practical scenarios where annotated biomedical data are scarce. Model performance is compared with state-of-the-art baselines, including pretrained vision-language models adapted with conventional fine-tuning strategies. Additionally, we perform ablation studies to quantify the contributions of each key component, CPFT, MGCL, and DSG. We further analyze how these components work together to enhance feature adaptation, cross-modal alignment, and class-level discriminability.

\tipparagraph{Datasets.}  
To ensure a comprehensive evaluation of the MVSL framework, we adopt the same 11 diverse biomedical imaging datasets as used in BiomedCoOp, spanning 10 anatomical regions and 9 distinct imaging modalities. Specifically, the datasets include: Computerized Tomography (CTKidney \cite{ctkidney}), Dermatoscopy (DermaMNIST \cite{dermamnist1,dermamnist2}), Endoscopy (Kvasir \cite{kvasir}), Fundus Photography (RETINA \cite{retina1,retina2}), Histopathology (LC25000 \cite{LC25000}, CHMNIST \cite{chmnist}), Magnetic Resonance Imaging (BTMRI \cite{btmri}), Optical Coherence Tomography (OCTMNIST \cite{octmnist}), Ultrasound (BUSI \cite{busi}), and X-Ray imaging (COVID-QU-Ex \cite{covid}, KneeXray \cite{kneexray}). These datasets collectively encompass a wide range of visual and anatomical variability, including diagnostically challenging modalities such as brain MRI and ultrasound. Such diversity enables a rigorous assessment of MVSL's robustness, generalizability, and effectiveness in few-shot medical image classification. Comprehensive task configurations and dataset partition details are presented in the \textit{Supplementary Materials}.

\tipparagraph{Few-Shot Learning.}  
To evaluate the model's ability to generalize under limited supervision, we conduct few-shot experiments with varying numbers of labeled examples per class ($K=1, 2, 4, 8, 16$). This setting simulates practical medical scenarios where expert-labeled data are often scarce, requiring high adaptation ability with minimal supervision.
 
\tipparagraph{Base-to-Novel Class Generalization.}  To evaluate the generalization ability of MVSL, each dataset is divided into base and novel class subsets. The model is trained on the base classes using a 16-shot setup and then evaluated on both base and novel classes. Performance is quantified using the harmonic mean (HM) of accuracy across base and novel classes, measuring the model’s capability to recognize unseen disease categories while retaining knowledge of seen classes. This setting mirrors real-world clinical scenarios, where reliable recognition of emerging or rare disease categories is critically required.

\tipparagraph{Experimental Configurations.}  
We follow the training strategy outlined in BiomedCoOp to implement the MVSL framework. Specifically, BiomedCLIP with a ViT-B/16 backbone is employed as the visual encoder, and all results are reported as the average of three independent runs to ensure statistical robustness. For few-shot classification, models are trained for 100 epochs, while base-to-novel generalization is evaluated after 50 training epochs. The learnable prompt context is initialized with the embedding of the phrase ``a photo of a'', and 50 textual prompts are generated using large language models. Further details are provided in the \textit{Supplementary Material}. Model optimization is carried out using stochastic gradient descent with a learning rate of 0.0025 and a batch size of 4. 

For fair comparison and stable optimization, the hyperparameters $\lambda_1$ and $\lambda_2$ are set following BiomedCoOp. In our framework, the newly introduced local losses maintain the same functional nature as their corresponding global counterparts. Accordingly, the local loss $\mathcal{L}_{\text{local}}$ is assigned the same weight as the global loss $\mathcal{L}_{\text{global}}$, which is set to $1.0$. Furthermore, both the DSG loss $\mathcal{L}_{\text{DSG}}$ and the MSE loss $\mathcal{L}_{\text{MSE}}$ impose constraints on the text representations, given their shared objective of regularizing the textual feature space, we apply identical weighting to these two terms and set $\lambda_1 = \lambda_3$. Detailed hyperparameter settings are provided in the \textit{Supplementary Material}. This design ensures consistent scaling across semantically related loss components while preserving comparability with prior work. In addition, the fusion coefficients $\beta_1$ and $\beta_2$ are defined as learnable parameters and initialized to 0.5. All experiments are conducted on a single NVIDIA RTX 4090 GPU with 24 GB of memory.

%%%%%%%%%%%%%%%%%%%%%%%%%
\begin{table*}[ht]
\centering
\caption{\textbf{Evaluation against state-of-the-art techniques.} The table summarizes the average classification accuracy (\%) achieved on 11 benchmark datasets. Each result is reported as mean, calculated over three randomly sampled support sets per dataset. The best-performing results are marked in \textbf{bold}, and the second-best results are denoted with an \underline{underline}. The red upward arrows and the numbers in parentheses indicate the performance gain over the second-best results.}
\tablestyle{-7pt}{1.1}
\addtolength{\tabcolsep}{+20pt}
\resizebox{\textwidth}{!}{%
\begin{tabular}{lccccccc}
\toprule
\textbf{Method} & Year& $K=1$ & $K=2$  & $K=4$  & $K=8$  & $K=16$ \\
\midrule
Linear Probing \cite{radford2021learning} &2021 &$48.91$  & $55.82$  & $62.12$ & $67.33$  & $70.81$\\
CoOp \cite{CoOp} & 2022& $52.59$ & $55.71$  &  $61.35$  & $67.74$ &  $71.48$ \\
CoCoOp \cite{cocoop} &2022 &$50.88$  &  $53.91$  &  $57.63$ &  $63.15$  & $67.51$   \\
Tip-Adapter \cite{zhang2021tip} & 2022&$50.35$ & $53.50$ & $58.33$  & $62.01$   & $67.60$ \\
Tip-Adapter-F \cite{zhang2021tip} &2022 &$52.55$ & $54.17$ & $62.30$  & $68.12$ & $68.12$ \\
ProGrad \cite{ProGrad} & 2023&$53.67$  & $56.42$  & $62.10$  & $67.06$  & $69.21$  \\
KgCoOp \cite{KgCoOp} &2023 &$54.31$  & $55.79$  & $60.92$  & $66.00$ & $67.71$ \\
CLIP-Adapter \cite{clip_adapter} &2023 &$45.53$ & $44.70$  & $45.30$  & $46.54$  & $48.46$  \\
MaPLe \cite{khattak2023maple} & 2023&$37.99$ & $40.89$ &  $44.09$ & $47.37$ & $52.93$ \\
PromptSRC \cite{PromptSRC} &2023&$50.72$  & $52.24$ & $56.46$  & $60.52$  & $63.22$  \\
LP++ \cite{huang2024lp++} & 2024&$49.27$ & $55.88$  & $61.30$ & $65.48$ & $70.09$ \\
GDA \cite{wang2024hard} &2024 &$49.56$ & $58.39$ & $63.41$  & $70.60$ & $72.86$ \\
XCoOp \cite{bie2024xcoop} & 2024&$52.50$ & $55.39$ & $60.87$  & $66.37$ & $71.04$ \\
DCPL \cite{cao2024domain} & 2024&$49.65$ & $58.65$ & $62.62$  & $68.65$ & $70.79$ \\
CLIP-LoRA \cite{zanella2024low} &2024& $48.31$ & $57.63$ & $62.31$  & $68.16$ & $70.31$ \\
ProKeR \cite{bendou2025proker} & 2025&$49.40$ & $58.84$ & $63.72$  & $70.98$ & $71.86$ \\
BiomedCoOp \cite{koleilat2025biomedcoop} &2025 &$\underline{56.87}$ & $\underline{59.32}$ & $\underline{64.34}$ & $\underline{68.96}$  & $\underline{73.41}$ \\
\midrule
 \textbf{MVSL (Ours)} &2025 & \textbf{58.41}~\textcolor{red}{$\uparrow$(1.38)}& \textbf{62.17} ~\textcolor{red}{$\uparrow$(3.04)} & \textbf{68.08}~\textcolor{red}{$\uparrow$(4.13)} & \textbf{73.10} ~\textcolor{red}{$\uparrow$(4.78)}& \textbf{77.13} ~\textcolor{red}{$\uparrow$(4.71)} \\
\bottomrule
\end{tabular}
}
\label{table:fewshot-main-results}
\end{table*}

\begin{figure}[!t]
\centering
    \begin{minipage}   {0.99\linewidth}
        \centering
        \includegraphics [width=1\linewidth] 
        {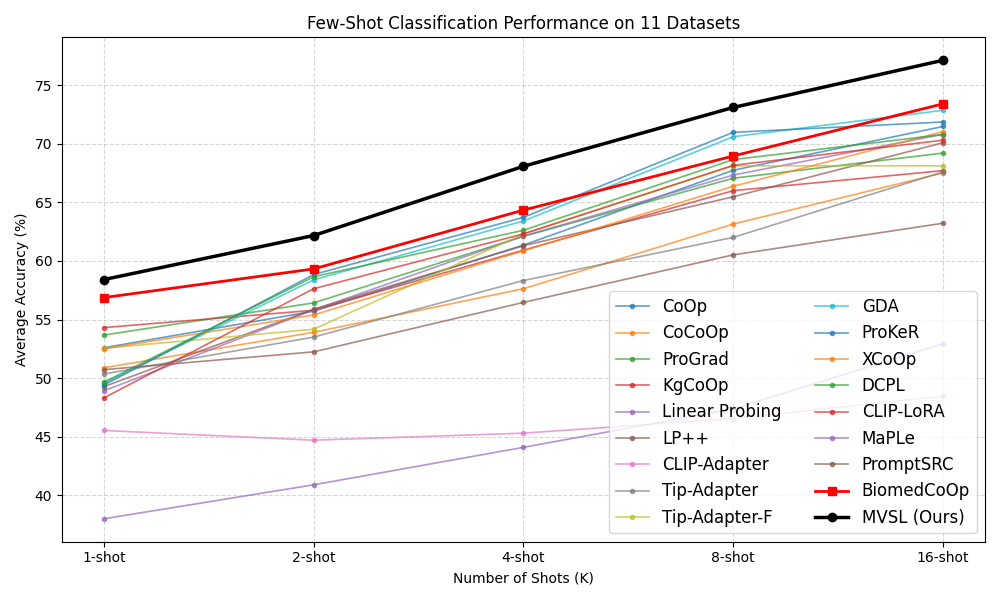}
    \end{minipage}
    \vspace{-0.2cm}
    \caption{Average classification accuracy (\%) of various few-shot adaptation methods across different numbers of training shots per class.}
    \vspace{-0.2cm}
    \label{fig:averageMVSL}
\end{figure}
%%%%%%%%%%%%%%%%%%%%%

%%%%%%%%%%%%%%%%%%%%%
\begin{table}[h]
\centering
\vspace{-0.2cm}
 \caption{\small\textnormal{Average accuracy comparison (\%) on Base-to-novel generalization of MVSL with other methods, computed across 11 datasets}.}
    \label{tab:base-to-new}
\tablestyle{1pt}{1.1}
\addtolength{\tabcolsep}{1pt}
\resizebox{\columnwidth}{!}{%
\begin{tabular}{lc|c c c c c c c}
\hline
\multirow{1}{*}{Dataset} &  &  {CoOp} & {CoCoOp} &  {KgCoOp} & 
{ProGrad} &{PromptSRC} &
{BiomedCoOp} & {\textbf{MVSL}} \\
\midrule
\multirow{3}{*}{\shortstack[l]{Average on\\  11 datasets}}       
& Base  & 73.85   & 72.26  & 68.36 & 71.67 & 69.01&76.29 & \textbf{80.05} \\
& Novel  & 64.75   & 67.03  & 64.08    & 66.93&70.99 &\textbf{76.29} & 74.58 \\
 & HM & 67.23 & 67.22  & 64.61  & 67.43  &69.99& 76.29 & \textbf{77.22} \\
    \bottomrule
    \end{tabular}
    }
    \vspace{-0.2cm}
\end{table}
%%%%%%%%%%%%%%%%%%%%%
%%%%%%%%%%%%%%%%%
%%%%%%%%%%%%%%%%%%%%%%%%%%%
\begin{table}[htbp]
\caption{Effect of different fine-tuning paradigm on few-shot and Base-to-Novel benchmarks.}
\label{tab:fine-tuning}
\centering
\tablestyle{-13pt}{1.1}
\arrayrulecolor{black}
\setlength\arrayrulewidth{1pt}
\addtolength{\tabcolsep}{+16pt}
\resizebox{\columnwidth}{!}{%
\begin{tabular}{lcccccccc}
\toprule
\multirow{2}{*}{Fine-tuning} & \multicolumn{3}{c}{Base-to-Novel} & \multicolumn{5}{c}{Few-shot} \\
\cmidrule(lr){2-4} \cmidrule(lr){5-9}
 & Base & Novel & HM & 1 & 2 & 4 & 8 & 16  \\
\midrule
$P_{\text{text}}$ & 75.94& 74.11 & 75.01& 57.01& 59.32 &64.44 & 69.12 & 72.46   \\
$P_{\text{img}}$ &63.58&60.21 & 61.85&49.45 & 51.02 &55.04& 57.39 & 58.85  \\   
$A_{\text{text}}$ &78.59 &52.30 & 62.80&53.21& 59.58 &65.56& 70.74 & 74.67  \\
$A_{\text{img}}$ &79.28&59.23 & 67.80& 55.88&60.17  &67.73&72.96  & \textbf{77.44}  \\
$P_{\text{text}}+P_{\text{img}}$ &75.92& 67.70& 71.57& 56.63& 59.39 &65.02& 69.25 &73.05   \\
$A_{\text{text}}+A_{\text{img}}$ &79.85&53.89 & 64.35& 55.03& 61.12 &65.97& 72.38 & 76.74  \\
$A_{\text{text}}+P_{\text{img}}$ &78.37&58.77 &67.17 & 54.43&59.89  &65.04& 70.99 & 74.54  \\
$P_{\text{text}}+A_{\text{img}}$ & \textbf{80.05}&\textbf{74.58}&\textbf{77.22} &\textbf{58.41}&\textbf{62.17}&\textbf{68.08}  &\textbf{73.10} &   77.13    \\
\bottomrule
\end{tabular}
}
\vspace{-0.2cm}
\end{table}
%%%%%%%%%%%%%%%%%%%%%%%%%%%

\subsection{Comparison with State-of-the-arts}
\tipparagraph{Few-Shot Evaluation}
As illustrated in Figure~\ref{fig:averageMVSL}, MVSL consistently delivers state-of-the-art performance across all few-shot settings, outperforming both prompt-based and adapter-based parameter-efficient baselines on 11 biomedical image benchmarks.  In particular, as reported in Table~\ref{table:fewshot-main-results}, under the most challenging low-shot regimes, MVSL achieves notable improvements over the strongest competitor, BiomedCoOp, with gains of 1.38\% at $K{=}1$ and 3.04\% at $K{=}2$. As the number of labeled samples increases, the performance gap widens further: MVSL surpasses BiomedCoOp by 4.13\%, 4.78\%, and 4.71\% at $K{=}4$, $K{=}8$, and $K{=}16$, respectively. Detailed few-shot evaluation results for each dataset are provided in the \textit{Supplementary Material}. These results demonstrate that MVSL scales effectively with increasing supervision, exhibiting both strong few-shot adaptability. 

It is worth noting that many parameter-efficient fine-tuning methods tend to plateau, or even exhibit diminishing returns, once the shot number exceeds $K{=}4$, suggesting limited ability to leverage additional supervision. In contrast, MVSL continues to deliver consistent performance gains across all sample sizes. This indicates that its multi-view collaborative mechanism can effectively exploit richer global–local and vision–language complementary cues when more labeled data are available, thereby capturing more informative cross-modal semantic relationships and enhancing generalization. Overall, MVSL achieves an excellent balance between accuracy, stability, and generalization across diverse few-shot scenarios. 

\tipparagraph{Base-to-Novel Generalization}
We evaluate the Base-to-Novel generalization of MVSL on 11 datasets and compare it with several existing methods, as shown in Table~\ref{tab:base-to-new}. On the base classes, MVSL achieves an average accuracy of 80.05\%, significantly outperforming BiomedCoOp (76.29\%) and other methods, demonstrating its strong ability to learn from seen categories. On the novel classes, MVSL attains an average accuracy of 74.58\%, slightly lower than BiomedCoOp (76.29\%) but still considerably higher than the other baselines, indicating good generalization to unseen categories. The harmonic mean (HM) further highlights that MVSL achieves 77.22\%, surpassing all compared methods, which shows that it maintains strong performance on base classes while achieving robust recognition of novel classes, demonstrating excellent Base-to-Novel generalization. Detailed base-to-novel evaluation results for each dataset are provided in the \textit{Supplementary Material}. 

\subsection{Ablation Study}
In this section, we present a comprehensive ablation study to systematically evaluate the individual contributions of the three core components of MVSL: Cross-Paradigm Fine-Tuning (CPFT), Multi-Granularity Contrastive Learning (MGCL), and the Disease Semantic Graph (DSG). Experiments are conducted on 11 diverse biomedical imaging datasets to rigorously assess how each component impacts the overall performance and generalization capability of the proposed framework.

\tipparagraph{The effects of CPFT.}
We conduct a comprehensive ablation study across different fine-tuning paradigms on both Base-to-Novel and few-shot benchmarks, as summarized in Table~\ref{tab:fine-tuning}. For single-branch tuning, applying prompt tuning to the text encoder ($P_{\text{text}}$) or adapter tuning to the image encoder ($A_{\text{img}}$) yields moderate gains on specific metrics, but their overall performance remains limited. Specifically, $P_{\text{text}}$ performs slightly better on Novel categories and low-shot settings, whereas $A_{\text{img}}$ excels on Base categories and high-shot settings.

When the same fine-tuning strategy is applied to both branches ($P_{\text{text}}+P_{\text{img}}$ or $A_{\text{text}}+A_{\text{img}}$), there is only minor improvement in certain Few-shot settings, and performance on Base-to-Novel HM still falls short of cross-paradigm combinations. This indicates that simply stacking identical strategies cannot fully leverage the complementary strengths of the text and visual branches.

In contrast, the cross-paradigm configuration $P_{\text{text}}+A_{\text{img}}$, applying prompt tuning to the text branch and adapter tuning to the image branch, achieves the best overall performance on Base-to-Novel HM and most Few-shot settings. This configuration effectively exploits the semantic control of text prompts and the visual representation power of image adapters, resulting in more stable cross-modal alignment and stronger low-shot generalization. 

Furthermore, another cross-paradigm setup, $A_{\text{text}}+P_{\text{img}}$, performs noticeably worse than $P_{\text{text}}+A_{\text{img}}$, further confirming that applying prompt tuning to the text branch and adapter tuning to the image branch is more advantageous in low-resource and cross-modal scenarios. Overall, these results support the design rationale of CPFT, which leverages cross-paradigm fine-tuning to achieve complementary synergy between text and visual branches, thereby enhancing few-shot learning performance.

%%%%%%%%%%%%%%%%%%%%%%%%%%%
\iffalse
\begin{table}[htbp]
\caption{Impact of different blocks selected in Residual Adapters on few-shot and Base-to-Novel benchmarks.}
\label{tab: block_selected}
\centering
\tablestyle{-13pt}{1.1}
\arrayrulecolor{black}
\setlength\arrayrulewidth{1pt}
\addtolength{\tabcolsep}{+16pt}
\resizebox{\columnwidth}{!}{%
\begin{tabular}{lcccccccc}
\toprule
\multirow{2}{*}{Blocks} & \multicolumn{3}{c}{Base-to-Novel} & \multicolumn{5}{c}{Few-shot} \\
\cmidrule(lr){2-4} \cmidrule(lr){5-9}
 & Base & Novel & HM & 1 & 2 & 4 & 8 & 16  \\
\midrule
%9& 79.50& 62.34&69.88&55.38 & 61.17 & 67.42& 73.37 & 77.95      \\
10 & 80.32& 71.60& 75.70&57.61 & 61.70 & \textbf{68.39}& 72.89 & 76.96        \\
11 &80.05 &\textbf{74.58}&\textbf{77.22}&58.41&62.17&68.08 &\textbf{73.10} & \textbf{77.13}    \\
12 & 78.96& 71.84&  75.23& 57.64 &61.50&67.12 &72.19  & 76.39      \\
10$+$11&79.85&72.23   &75.85 &58.48 &\textbf{62.72}&68.17& 72.97& 76.47\\
10$+$12 &79.72& 66.92& 72.76& \textbf{58.99}& 62.46 &67.48& 72.75 & 76.62  \\
11$+$12 &79.96& 65.11& 71.77& 58.69&61.97  &68.15& 72.92 &76.90   \\
10$+$11$+$12&\textbf{80.36}&71.48&75.66 &56.89&61.64&67.00  &72.32 &    76.15    \\
\bottomrule
\end{tabular}
}
\end{table}
\fi
%%%%%%%%%%%%%%%%%%%%%
%%%%%%%%%%%%%%%%%%%%%
\begin{figure*}[!t]
\centering
    \begin{minipage}   {0.99\linewidth}
        \centering
        \includegraphics [width=1\linewidth] 
        {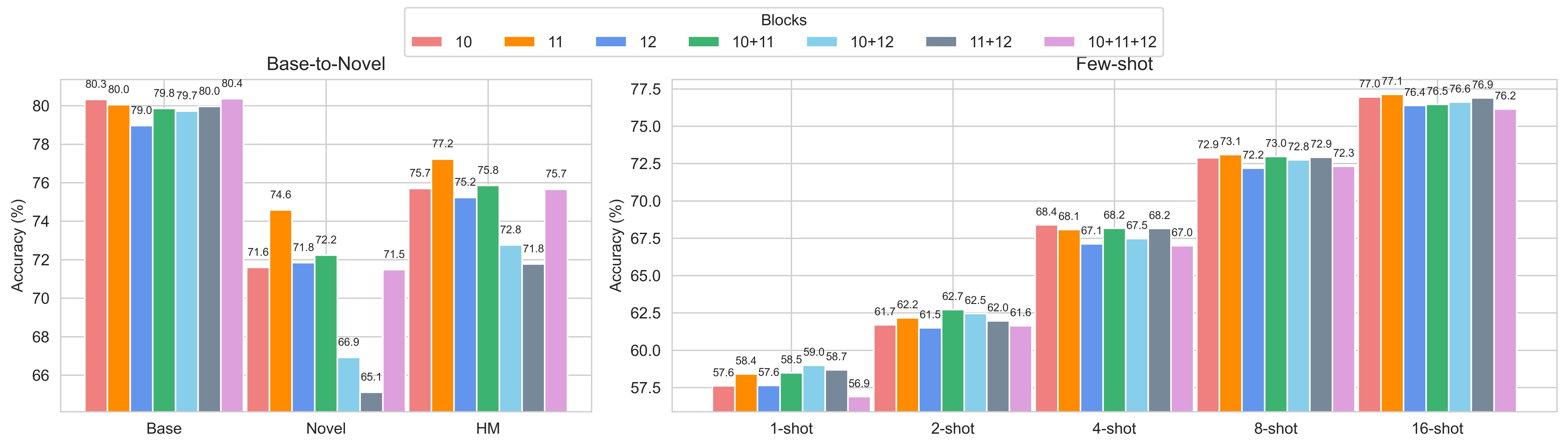}
    \end{minipage}
    \vspace{-0.2cm}
    \caption{Impact of inserting visual residual adapters (VRA) into the last three Transformer blocks of the ViT backbone on Base-to-Novel and few-shot performance. Each block represents a different combination of residual adapters. The left subfigure shows Base-to-Novel metrics (Base, Novel, HM), and the right subfigure shows few-shot recognition accuracy across 1-, 2-, 4-, 8-, and 16-shot settings. Block $11$ demonstrates the most significant improvement, highlighting the effectiveness of placing adapters in the upper layers, whereas adding adapters to multiple blocks does not guarantee linear gains and may even degrade generalization. Colors correspond to different block configurations as indicated in the legend.}
    \vspace{-0.2cm}
    \label{fig:block}
\end{figure*}
%%%%%%%%%%%%%%%%%%%%%
\tipparagraph{Vision residual adapter is essential but could also become evil.}
Following the design insights from CLIP-Adapter, we focus on inserting residual-style adapters primarily into the higher layers of the ViT backbone rather than all layers. CLIP-Adapter demonstrates that placing adapters in the upper layers of the visual encoder achieves optimal performance with minimal computational cost, whereas inserting adapters in earlier layers not only increases the computational burden for gradient back-propagation but may also interfere with the pretrained knowledge in CLIP. Inserting adapters into all layers substantially increases the number of trainable parameters, which can easily lead to overfitting on few-shot data. Therefore, in our ablation studies, we restrict the placement of residual adapters to the last few Transformer blocks to balance parameter efficiency, preservation of pretrained knowledge, and few-shot generalization.

To illustrate the differences among blocks, we provide a bar chart visualization of the Base-to-Novel and few-shot results in Figure~\ref{fig:block}, where each bar color corresponds to a different block configuration. The results indicate that block $11$ is the most critical: adding VRA solely to block $11$ achieves the best or near-best results on Base-to-Novel and most few-shot settings, suggesting that this layer provides the most effective balance between cross-domain adaptation and discriminative feature capture. In contrast, modifying only block $10$ or block $12$ yields limited gains.

Moreover, combining multiple blocks does not result in linear improvements and can even impair generalization. For instance, while adding VRA to blocks $10+11$ or $10+12$ provides slight gains in certain few-shot settings, the overall Base-to-Novel performance remains below that of block $11$ alone; when all three blocks $10+11+12$ are equipped with adapters, although performance on the Base set slightly increases, Novel and HM drop significantly. This indicates that, although visual residual adapters are necessary, excessive stacking can lead to representation drift and hurt the model’s adaptability to novel classes.

In summary, block $11$ is the most representative placement, effectively enhancing few-shot recognition and cross-domain generalization, whereas unnecessary multi-layer insertion may introduce overfitting or disrupt global feature consistency.

\tipparagraph{The benefits of fixed fusion coefficient $\alpha$ in VRA.}
As shown in Table~\ref{tab:fusion coefficient}, the fusion coefficient $\alpha$ in the VRA plays a pivotal role in governing the degree to which residual updates influence the pretrained visual backbone. Extremely small values (e.g., $\alpha = 0.0$ or $0.1$) significantly underutilize the adapter branch, leading to weak feature adaptation and consequently poor performance across both Base-to-Novel and few-shot benchmarks. Conversely, large fusion weights (e.g., $\alpha = 0.9$ or $1.0$) cause the adapter to dominate the fused representation, which disrupts the pretrained visual structure and results in decreased generalization, especially in low-shot settings.

Performance peaks at $\alpha = 0.5$, consistently achieving the highest accuracy in both benchmarks (e.g., Base-to-Novel HM of $77.22\%$ and strong improvements across 1-, 4-, 8-, and 16-shot settings). This suggests that an equal blend between pretrained features and adapter-induced updates yields the most effective trade-off between stability and task-specific adaptability. 

\begin{figure}[!t]
\centering
    \begin{minipage}   {0.99\linewidth}
        \centering
        \includegraphics [width=1\linewidth] 
        {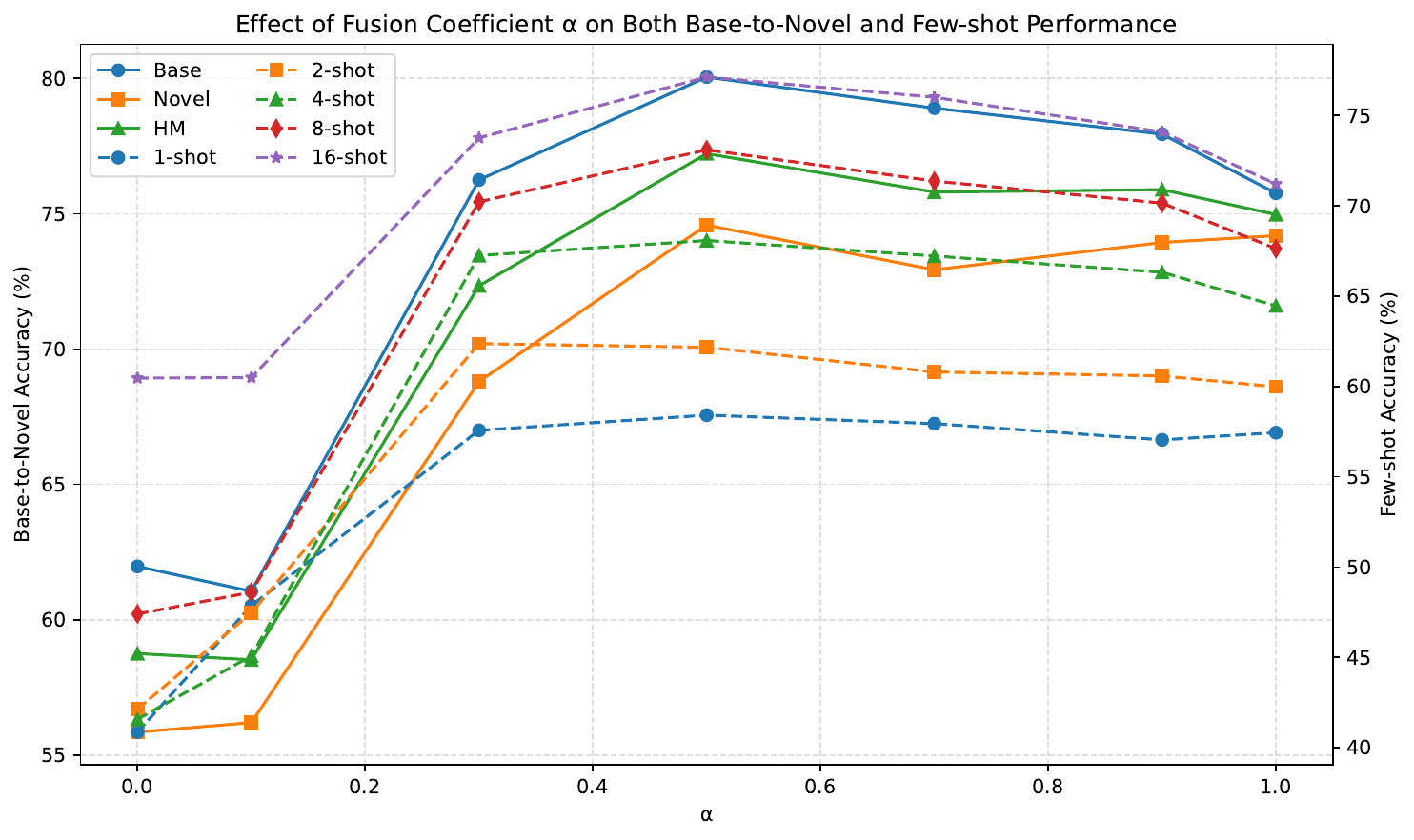}
    \end{minipage}
    \vspace{-0.2cm}
    \caption{Effect of the fusion coefficient $\alpha$ on Base-to-Novel and Few-shot performance. The curves show Base-to-Novel metrics (Base, Novel, HM) and Few-shot accuracies (1-, 2-, 4-, 8-, and 16-shot) as $\alpha$ varies from 0.0 to 1.0. Peak performance is observed at intermediate $\alpha$ values (around 0.5), indicating that a balanced fusion of local and global features optimally enhances classification performance. This figure highlights the sensitivity of the model to $\alpha$ and provides guidance for selecting an effective fusion setting.}
    \vspace{-0.2cm}
    \label{fig:fusion_alpha}
\end{figure}

Interestingly, the learnable fusion coefficient $\alpha^*$, though capable of adjusting dynamically during training, does not surpass the fixed $\alpha = 0.5$ configuration. The observed suboptimal performance of $\alpha^*$ indicates that allowing $\alpha$ to vary freely may introduce instability, especially when training data is scarce, leading to overfitting or excessive reliance on the adapter pathway. As shown in Figure~\ref{fig:fusion_alpha}, the line-plot visualization further confirms that $\alpha = 0.5$ lies at the center of the optimal performance plateau.

Overall, the ablation indicates that a fixed, moderate fusion coefficient provides more stable and reliable structural adaptation, validating the use of $\alpha = 0.5$ as the default setting in VRA.

%%%%%%%%%%%%%%%%%%%%%%%%%%%
\begin{table}[htbp]
\caption{Impact of varying fusion coefficients in Visual Residual Adapters on few-shot and Base-to-Novel benchmarks, where $\alpha^*$ is defined as a learnable parameter (initialized as 0.5) that represents the fusion weight between different feature components in the model.}
\label{tab:fusion coefficient}
\centering
\tablestyle{-13pt}{1.1}
\arrayrulecolor{black}
\setlength\arrayrulewidth{1pt}
\addtolength{\tabcolsep}{+16pt}
\resizebox{\columnwidth}{!}{%
\begin{tabular}{lcccccccc}
\toprule
\multirow{2}{*}{$\alpha$} & \multicolumn{3}{c}{Base-to-Novel} & \multicolumn{5}{c}{Few-shot} \\
\cmidrule(lr){2-4} \cmidrule(lr){5-9}
 & Base & Novel & HM & 1 & 2 & 4 & 8 & 16  \\
\midrule
0.0 &61.97 &55.85 & 58.75 &40.86 &42.13 & 41.57 & 47.40 & 60.46    \\
0.1& 61.05& 56.20 &58.52 &47.86 &47.47 & 45.08 & 48.60  &   60.49       \\
0.3 &76.25&68.80 & 72.34&57.57& \textbf{62.37}&67.25&70.22  &  73.76     \\
0.5&\textbf{80.05} &\textbf{74.58}&\textbf{77.22}&\textbf{58.41} &62.17&\textbf{68.08} &\textbf{73.10} & \textbf{77.13}    \\
0.7&78.90&72.93&75.80&57.94&60.80&67.22&71.37&76.01 \\
0.9&77.94&73.94&75.89&57.04& 60.58&66.32&70.15 &74.09       \\
1.0& 75.76& 74.19 &74.97 &57.44 &59.98  & 64.46&67.63 &  71.21    \\
$\alpha^*$& 79.99 &68.83  & 73.99&57.47&62.06 & 67.83&72.32&  76.21     \\
\bottomrule
\end{tabular}
}
\vspace{-0.2cm}
\end{table}

%%%%%%%%%%%%%%%%%%%%%
%%%%%%%%%%%%%%%%%%%%%
%%%%%%%%%%%%%%%%%%%%%%%%%%%%%%%%%%%%%%%%%%%%
%%%%%%%%%%%%%%%%%%%%%%%%%%%
%\iffalse
\begin{table}[htbp]
\caption{Impact of MGCL and DSG adopted in MVSL on few-shot and Base-to-Novel benchmarks.}
\label{ablation:local}
\centering
\tablestyle{-13pt}{1.1}
\arrayrulecolor{black}
\setlength\arrayrulewidth{1pt}
\addtolength{\tabcolsep}{+16pt}
\resizebox{\columnwidth}{!}{%
\begin{tabular}{lcccccccc}
\toprule
\multirow{2}{*}{Prediction} & \multicolumn{3}{c}{Base-to-Novel} & \multicolumn{5}{c}{Few-shot} \\
\cmidrule(lr){2-4} \cmidrule(lr){5-9}
 & Base & Novel & HM & 1 & 2 & 4 & 8 & 16  \\
\midrule
$S_\text{global}$&80.08& 71.47 & 75.53& 56.89& 60.77 &67.16 & 72.89 & 77.02  \\
$S_\text{global} + S_\text{local}$ & 80.05 &74.58  & 77.22& 58.41&62.17    & 68.08 & 73.10  & 77.13  \\
\bottomrule
\end{tabular}
}
\vspace{-0.2cm}
\end{table}
%\fi
%%%%%%%%%%%%%%%%%%%%%%%%%%%
%%%%%%%%%%%%%%%%%%%%%%%%%%%

%%%%%%%%%%%%%%%%%%%%%%%%%%%
\begin{table}[htbp]
\caption{Impact of MGCL and DSG adopted in MVSL on few-shot and Base-to-Novel benchmarks.}
\label{ablation:modules}
\centering
\tablestyle{-13pt}{1.1}
\arrayrulecolor{black}
\setlength\arrayrulewidth{1pt}
\addtolength{\tabcolsep}{+16pt}
\resizebox{\columnwidth}{!}{%
\begin{tabular}{lcccccccc}
\toprule
\multirow{2}{*}{Components} & \multicolumn{3}{c}{Base-to-Novel} & \multicolumn{5}{c}{Few-shot} \\
\cmidrule(lr){2-4} \cmidrule(lr){5-9}
 & Base & Novel & HM & 1 & 2 & 4 & 8 & 16  \\
\midrule
Baseline& 75.94& 74.11 & 75.01& 57.01& 59.32 &64.44 & 69.12 & 72.46  \\
+ MGCL& \textbf{80.66} & 69.69  & 74.78 & 57.69 & 61.71 & 67.88&  \textbf{73.89} &  \textbf{77.81} \\
+ DSG&75.76& 74.19 &74.97 &57.44 &59.98 &64.46&67.63 & 71.21   \\
+ MGCL + DSG & 80.05 & \textbf{74.58}  & \textbf{77.22} & \textbf{58.41}&  \textbf{62.17}     & \textbf{68.08}& 73.10  & 77.13  \\
\bottomrule
\end{tabular}
}
\vspace{-0.2cm}
\end{table}
%%%%%%%%%%%%%%%%%%%%%
\begin{figure}[!t]
\centering
    \begin{minipage}   {1.0\linewidth}
        \centering
        \includegraphics [width=1\linewidth] 
        {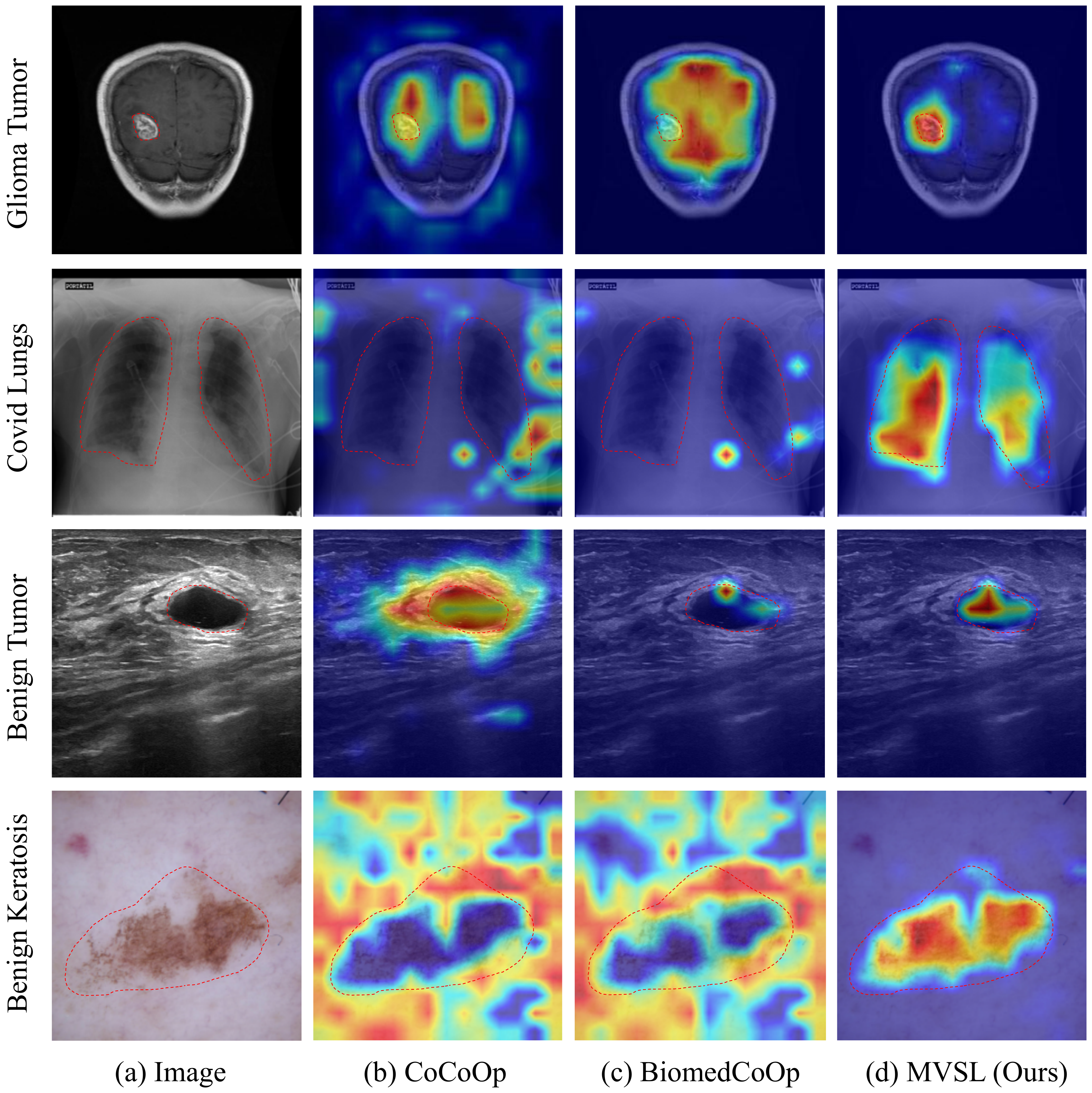}
    \end{minipage}
    \vspace{-0.2cm}
   \caption{Visual saliency maps across various medical imaging modalities. Each row corresponds to a specific disease category, and each column compares the original image, CoCoOp, BiomedCoOp, and the proposed MVSL. MVSL consistently yields more concentrated and clinically aligned attention within the annotated lesion regions (red contours), demonstrating superior visual interpretability. }
    \label{fig:vis}
    \vspace{-0.2cm}
\end{figure}
%%%%%%%%%%%%%%%%%%%%%

\tipparagraph{MGCL and DSG are mutually essential.}
As shown in Table~\ref{ablation:modules}, MGCL and DSG provide complementary benefits, and integrating both components yields the strongest overall performance. Introducing MGCL alone substantially boosts the Base accuracy from 75.94\% to 80.66\% and improves mid- to high-shot performance (4-, 8-, 16-shots), demonstrating that multi-granularity alignment effectively enhances visual–text correspondence and strengthens feature robustness when more visual evidence is available. However, MGCL slightly decreases Novel accuracy in the Base-to-Novel evaluation (from 74.11\% to 69.69\%), indicating that local alignment alone may overfit base categories without relational regularization.

In contrast, DSG alone delivers a balanced improvement on Novel categories (from 74.11\% to 74.19\%) and slightly stabilizes few-shot performance, confirming that disease-level semantic priors help regulate inter-class relationships and prevent representation collapse. However, DSG does not significantly improve Base accuracy or performance under higher shots, showing that semantic regularization alone is insufficient for refining cross-modal alignment.

When MGCL and DSG are combined, MVSL achieves the best overall results across both benchmarks. Base-to-Novel HM increases from 75.01\% to 77.22\%, and few-shot accuracy improves consistently in 1-, 2-, and 4-shot settings, demonstrating that DSG offsets MGCL’s tendency to overfit, while MGCL provides the fine-grained discriminability absent in DSG. This complementary interaction confirms that MGCL and DSG address different aspects of the low-resource classification challenge, fine-grained alignment and semantic structural regularization, and are mutually essential for unlocking the full potential of MVSL.
%%%%%%%%%%%%%%%%%%%%%
\begin{figure*}[!t]
\centering
    \begin{minipage}   {1.0\linewidth}
        \centering
        \includegraphics [width=1\linewidth] 
        {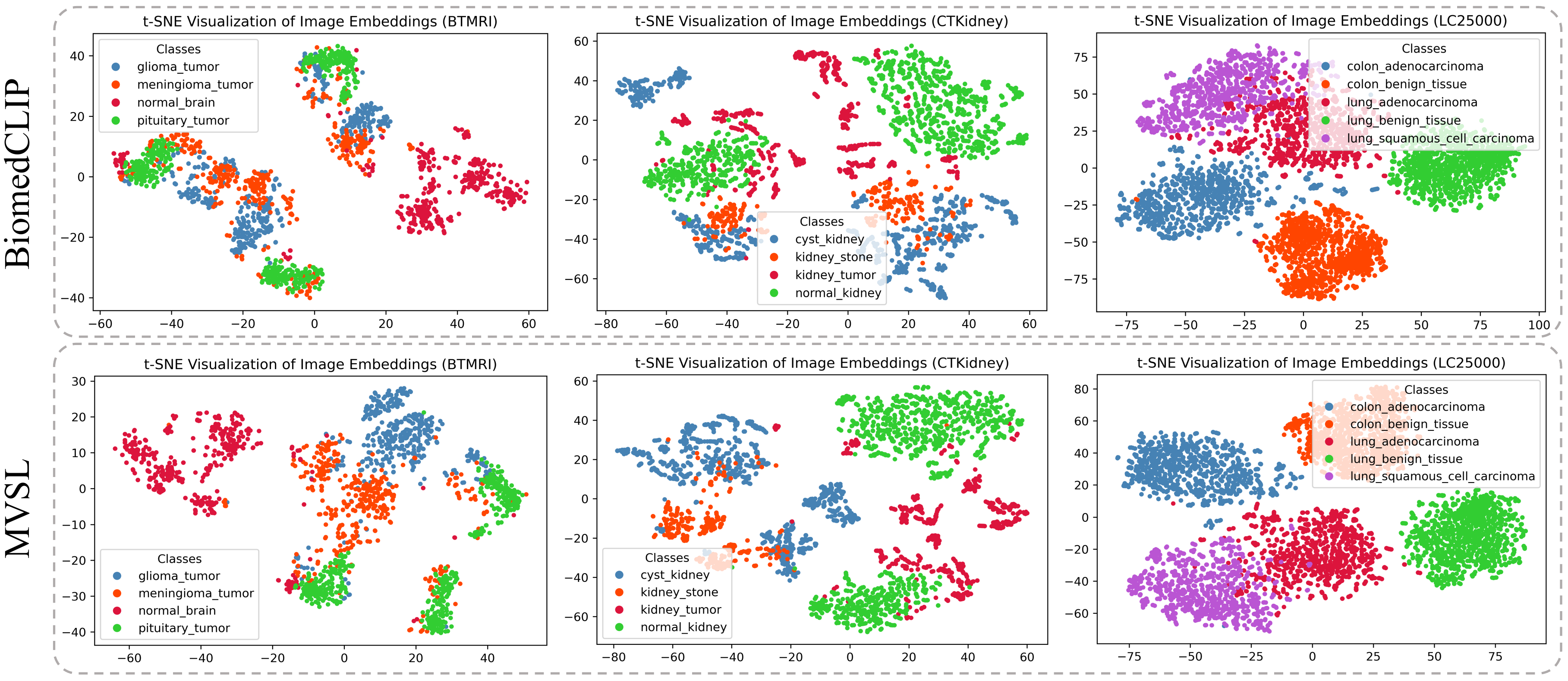}
    \end{minipage}
    \vspace{-0.2cm}
   \caption{t-SNE visualization of image embeddings generated by MVSL compared with pretrained BiomedCLIP. The embeddings from MVSL form more compact and well-separated clusters, indicating improved discriminability in the learned feature space. Experiments are conducted on BTMRI, CTKidney, and LC25000 datasets.}
\label{fig:tsne}
\vspace{-0.2cm}
\end{figure*}
%%%%%%%%%%%%%%%%%%%%%

\tipparagraph{Local predictions matters.}
To evaluate the contribution of our local prediction, we conduct an ablation study comparing the performance of using only global predictions $S_{\text{global}}$ versus combining both global and local predictions $S_{\text{global}} + S_{\text{local}}$. As shown in Table \ref{ablation:local}, introducing local predictions yields consistent improvements across both Base-to-Novel and few-shot benchmarks.

Specifically, the harmonic mean on the Base-to-Novel setting increases from $75.53\%$ to $77.22\%$, driven primarily by a notable improvement on the Novel classes ($+3.11\%$). This indicates that local predictions effectively complement the global classifier, helping the model better capture fine-grained visual cues that are critical for generalizing to unseen categories.

Similarly, under the few-shot setting, integrating local predictions brings consistent gains across all shot numbers. For example, the 1-shot accuracy rises from $56.89\%$ to $58.41\%$, with improvements maintained up to the 16-shot case. These results demonstrate that local prediction enhances the model’s ability to leverage limited training samples by refining feature discrimination at region-level granularity.

Overall, the empirical evidence confirms that local predictions play a crucial role in strengthening both cross-domain generalization and few-shot learning performance. Integrating them with global predictions leads to a more robust and complementary representation, validating the motivation behind our multi-view semantic learning framework.

\tipparagraph{Qualitative analysis.}
We further compare the visual interpretability of different approaches by using gScoreCAM~\cite{chen2022gscorecam}. As illustrated in Figure~\ref{fig:vis}, MVSL consistently produces more accurate and clinically meaningful saliency maps across diverse medical imaging modalities, including MRI (Glioma Tumor), X-ray (Covid Lungs), ultrasound (Benign Tumor), and dermoscopy (Benign Keratosis). Compared with CoCoOp and BiomedCoOp, our MVSL yields notably sharper and more localized activation regions that better align with the annotated lesion areas (highlighted by red contours). For instance, in both tumor-related cases (Glioma and Benign Tumor), MVSL concentrates its attention precisely on the pathological regions rather than spreading across irrelevant tissue. Similarly, in Covid Lungs and dermoscopic keratosis images, MVSL captures the essential disease-specific structures with clearer boundaries and reduced noise. These qualitative results demonstrate that MVSL not only improves classification accuracy but also enhances visual interpretability, offering more reliable and clinically aligned spatial attention, an essential property for trustworthy deployment in real-world medical image analysis.

In addition to spatial interpretability, we employed t-SNE~\cite{tsne} to visualize the feature embeddings generated by MVSL and compared them with those from the pretrained BiomedCLIP. As shown in Figure~\ref{fig:tsne}, MVSL produces more compact and well-separated clusters, indicating improved discriminability in the learned embedding space. This analysis is conducted on BTMRI, CTKidney, and LC25000 datasets. The observed clustering improvements suggest that MVSL effectively structures the feature space, complementing its enhanced visual interpretability and potentially supporting more robust downstream task performance.

\section{Conclusion}
We introduce MVSL, a unified Multi-View Synergistic Learning framework for accurate and generalizable biomedical image classification under low-resource conditions. Through cross-paradigm fine-tuning, MVSL decouples visual and textual adaptation: lightweight residual adapters refine visual features, while learnable prompts inject task-specific semantics into the textual branch, effectively stabilizing cross-modal alignment. Its multi-granularity contrastive learning captures both global and local image–text correspondences, and a dynamic fusion mechanism integrates these predictions to ensure robust inference. Furthermore, the disease semantic graph encodes inter-class relationships and serves as a teacher prior to guide textual branch fine-tuning via structural distillation, enhancing inter-class discriminability and indirectly improving visual embeddings. Extensive experiments on 11 public biomedical datasets spanning 9 imaging modalities and 10 anatomical regions demonstrate that MVSL consistently surpasses state-of-the-art methods in few-shot and zero-shot settings. These results confirm that MVSL, through cross-paradigm fine-tuning, multi-granularity alignment, and semantic-structural guidance, provides a scalable, interpretable, and robust framework for biomedical vision-language models, offering a promising solution for reliable, flexible, and clinically aligned multimodal understanding in low-resource scenarios.

{\small
\bibliographystyle{ieee_fullname}
\bibliography{egbib}

\clearpage
\section{Appendix}
\subsection{Dataset Details}
We evaluate our proposed method on 11 biomedical image classification datasets, covering 9 imaging modalities, including CT, MRI, X-ray, and ultrasound, and involving 10 different organs. For each dataset, we provide detailed specifications, including imaging modality, target organ(s), number of classes, and the splits for training, validation, and testing, as summarized in Table \ref{tab:dataset_info}.

These datasets encompass a wide range of clinically relevant scenarios, such as kidney cysts in CT images, various skin lesion categories in dermoscopic images, and multiple severity stages of knee osteoarthritis in X-ray images. The diversity of disease categories and imaging modalities allows for a comprehensive evaluation of model generalization across heterogeneous biomedical classification tasks, providing a representative and challenging benchmark.

Consistent with BiomedCoOp, we do not rely on the full training splits. Instead, we employ random few-shot sampling strategies to construct the training sets, enabling efficient and representative learning under limited-data conditions. Moreover, samples from each class are proportionally distributed across the training, validation, and test splits to ensure class balance. This experimental setup facilitates reliable evaluation on clinically relevant data and further enhances the robustness and fairness of performance comparisons across different datasets and tasks.
%%%%%%%%%%%%%%%%%%%%%
\begin{table*}[h]
\tablestyle{-27pt}{1.1}
\caption{A comprehensive overview of 11 datasets spanning 9 biomedical imaging modalities and 10 organs.}%
\label{tab:dataset_info}
\addtolength{\tabcolsep}{+30pt}
\resizebox{\textwidth}{!}{%
\begin{tabular}{|c|c|c|c|c|c|c|}
\hline
\textbf{Modality}                            & \textbf{Organ(s)}    & \textbf{Name} & \textbf{Classes}       & \textbf{Train}  &\textbf{Val}  & \textbf{Test}  \\ \hline
Computerized Tomography    & Kidney      & CTKidney    & \begin{tabular}[c]{@{}c@{}} Kidney Cyst, Kidney Stone, \\ Kidney Tumor, Normal Kidney\end{tabular} & 6221&2487&3738             \\ \hline
Dermatoscopy & Skin   & DermaMNIST  & \begin{tabular}[c]{@{}c@{}}Actinic Keratosis, Basal Cell Carcinoma, \\ Benign Keratosis, Dermatofibroma, \\ Melanocytic nevus, Melanoma, Vascular Lesion\end{tabular} & 7007&1003&2005 \\ \hline
Endoscopy & Colon & Kvasir  & \begin{tabular}[c]{@{}c@{}}Dyed Lifted Polyps, Normal Cecum, \\ Esophagitis, Dyed Resection Margins, \\ Normal Pylorus, Normal Z Line, \\ Polyps, Ulcerative Colitis\end{tabular}                 & 2000&800&1200              \\ \hline
Fundus Photography   & Retina   & RETINA     & \begin{tabular}[c]{@{}c@{}}Cataract, Diabetic Retinopathy, \\ Glaucoma, Normal Retina\end{tabular}  & 2108&841&1268                                                                                                            \\ \hline
\multirow{5}{*}{Histopathology}  & \begin{tabular}[c]{@{}c@{}}Lung\\ Colon\end{tabular} & LC25000    & \begin{tabular}[c]{@{}c@{}}Colon Adenocarcinoma, Colon Benign Tissue, \\ Lung Adenocarcinoma, Lung Benign Tissue, \\ Lung Squamous Cell Carcinoma\end{tabular}     & 12500&5000&7500           \\ \cline{2-7} 
    & Colorectal     & CHMNIST     & \begin{tabular}[c]{@{}c@{}}Adipose Tissue, Complex Stroma, \\ Debris, Empty Background, \\ Immune Cells, Normal Mucosal Glands, \\ Simple Stroma, Tumor Epithelium\end{tabular}                   & 2496&1000&1504             \\ \hline
\multirow{1}{*}{Magnetic Resonance Imaging} & \multirow{1}{*}{Brain}    & BTMRI   & \begin{tabular}[c]{@{}c@{}}Glioma Tumor, Meningioma Tumor, \\ Normal Brain, Pituitary Tumor\end{tabular}                                                                                          & 2854&1141&1717            \\ \hline
Optical Coherence Tomography               & Retina   & OCTMNIST  & \begin{tabular}[c]{@{}c@{}}Choroidal Neovascularization, Drusen, \\ Diabetic Macular Edema, Normal \end{tabular}                                                                                                & 97477&10832&1000   \\ \hline
Ultrasound                                   & Breast   & BUSI & \begin{tabular}[c]{@{}c@{}}Benign Tumors, Malignant Tumors, \\ Normal Scans\end{tabular}    & 389&155&236                \\ \hline                                                                                                      
\multirow{3}{*}{X-Ray}                       & Chest       & COVID-QU-Ex & \begin{tabular}[c]{@{}c@{}}COVID-19, Lung Opacity, \\ Normal Lungs, Viral Pneumonia\end{tabular}                                                                                                  & 10582&4232&6351           \\ \cline{2-7} 
    & Knee    & KneeXray & \begin{tabular}[c]{@{}c@{}}No, Doubtful, Minimal, \\ Moderate, and Severe Osteoarthritis\end{tabular}                                                                                             & 5778&826&1656              \\ \hline
\end{tabular}
}
\end{table*}
%%%%%%%%%%%%%%%%%%%%%

\subsection{Additional Hyperparameters}
Table~\ref{tab:hyperparameters} summarizes the hyperparameters ($\lambda_1$, $\lambda_2$, and $\lambda_3$) chosen for MVSL across various datasets in few-shot and base-to-novel benchmarks. These parameters were tuned to achieve a balance between classification accuracy and model adaptability. For fair comparison and stable optimization, the hyperparameters $\lambda_1$ and $\lambda_2$ follow the settings adopted in BiomedCoOp. Given that both the DSG loss $\mathcal{L}_{DSG}$ and the MSE loss $\mathcal{L}_{MSE}$ impose constraints on the text representations, given their shared objective of regularizing the textual feature space, we use an identical weighting for these two terms and set $\lambda_1 = \lambda_3$. 
%%%%%%%%%%%%%%%%%%%%%%%%%
\begin{table}[h]
\centering
\tablestyle{-12pt}{1.1}
\caption{\small\textnormal{Hyperparameter values for $\lambda_1$, $\lambda_2$, and $\lambda_3$, across different datasets and benchmarks.}}
\label{tab:hyperparameters}
\addtolength{\tabcolsep}{+20pt}
\resizebox{\columnwidth}{!}{%
\begin{tabular}{lc|c c c c c}
\hline
{Dataset} & Benchmark & {$\lambda_1$} &  {$\lambda_2$} &  {$\lambda_3$} \\
\midrule
\multirow{2}{*}{BTMRI}     & Few-shot & 0.5  & 0.25  & 0.5    \\
                           &  Base-to-Novel   &0.5  & 0.5 &0.5    \\
\midrule
\multirow{2}{*}{BUSI}     & Few-shot &0.75  & 0.75 &0.75    \\
                           &  Base-to-Novel   &0.5 & 0.5  & 0.5 \\
\midrule
\multirow{2}{*}{COVID-QU-Ex}     & Few-shot & 0.5  & 2.0  & 0.5  \\
                           &  Base-to-Novel  & 20.0  & 1.0 & 20.0  \\
\midrule
\multirow{2}{*}{CTKIDNEY}     & Few-shot & 1.0  & 0.5  & 1.0  \\
                           &  Base-to-Novel   & 10.0  & 0.25  & 10.0  \\
\midrule
\multirow{2}{*}{DermaMNIST}     & Few-shot & 5.0  & 20.0  & 5.0 \\
                           &  Base-to-Novel   & 2.0  & 0.5  & 2.0 \\
\midrule
\multirow{2}{*}{Kvasir}     & Few-shot & 0.75  & 0.75   & 0.75 \\
                           &  Base-to-Novel  & 1.0  & 1.0 & 1.0   \\
\midrule
\multirow{2}{*}{CHMNIST}     & Few-shot & 0.25  & 0.25 & 0.25   \\
                           &  Base-to-Novel  & 10.0  & 1.0  & 10.0  \\
\midrule
\multirow{2}{*}{LC25000}     & Few-shot & 0.5  & 0.5  & 0.5 \\
                           &  Base-to-Novel   & 0.25  & 0.75   & 0.25  \\
\midrule
\multirow{2}{*}{RETINA}     & Few-shot& 0.25  & 0.25  & 0.25  \\
                           &  Base-to-Novel  &5.0  & 1.0  &5.0  \\
\midrule
\multirow{2}{*}{KneeXray}     & Few-shot & 5.0  & 20.0 & 5.0  \\
                           &  Base-to-Novel  & 0.25  & 3.0 & 0.25  \\
\midrule
\multirow{2}{*}{OCTMNIST}     & Few-shot & 1.0  & 0.75 & 1.0  \\
                           &  Base-to-Novel   & 0.75  & 0.5  & 0.75 \\
\bottomrule
\end{tabular}%
}
%\vspace{-5mm}
\end{table}
%%%%%%%%%%%%%%%%%%%%%

\subsection{Adopted LLM Prompts}   
Following BiomedCoOp, we provide here a single text prompt generated by GPT-4 for each class across all datasets:
\\ \\
{
``\texttt{The image of a normal brain on MRI shows a clear differentiation between different brain regions with no disruptions.}"\\ \\
\noindent
``\texttt{Central necrosis and surrounding edema in glioma tumor on MRI scan.}"\\ \\
\noindent
``\texttt{Meningioma tumor on MRI displaying a dural tail sign and homogeneous enhancement.}"\\ \\
\noindent
``\texttt{Pituitary tumors often cause sellar expansion and may invade adjacent structures.}"\\ \\
\noindent
``\texttt{A routine ultrasound showing a hypoechoic, well-defined nodule, indicating a benign breast tumor.}"\\ \\
\noindent
``\texttt{An ultrasound revealing microcalcifications within the mass, indicating a malignant breast tumor.}"\\ \\
\noindent
``\texttt{A grayscale ultrasound highlighting well-defined ducts and lobules, characteristic of a normal breast ultrasound scan.}"\\ \\
\noindent
``\texttt{An X-ray scan showing bilateral airspace consolidation, typical of covid lungs.}"\\ \\
\noindent
``\texttt{A chest X-ray image with reticular and nodular opacities, indicative of lung opacity lungs.}"\\ \\
\noindent
``\texttt{An X-ray revealing no signs of consolidation or effusion, suggesting normal lungs.}"\\ \\
\noindent
``\texttt{An X-ray image revealing multifocal ground-glass and consolidative opacities, indicative of viral pneumonia lungs.}"\\ \\
\noindent
``\texttt{A CT image showing a lesion with uniform density and no internal irregularities, indicative of a cyst kidney.}"\\ \\
\noindent
``\texttt{A CT scan showing a calcified structure with acoustic shadowing, consistent with a kidney stone.}"\\ \\
\noindent
``\texttt{A CT scan showing a lesion with poorly defined margins, consistent with a kidney tumor.}"\\ \\
\noindent
``\texttt{A CT image revealing no signs of renal atrophy or cortical thinning, suggesting a normal kidney.}"\\ \\
\noindent
``\texttt{Actinic keratosis lesions may become thicker and more pronounced over time without treatment.}"\\ \\
\noindent
``\texttt{BCC lesions may bleed with minor trauma, such as shaving, due to their friable nature.}"\\ \\
\noindent
``\texttt{Cryotherapy, using liquid nitrogen, is a common treatment for seborrheic keratosis, causing the lesions to blister and fall off.}"\\ \\
\noindent
``\texttt{Dermatofibromas can be multiple in patients with systemic lupus erythematosus or other autoimmune conditions.}"\\ \\
\noindent
``\texttt{A clinical image with a lesion that has changed in size or texture, indicative of melanoma.}"\\ \\
\noindent
``\texttt{Melanocytic nevi can become darker and larger during pregnancy due to hormonal changes and increased melanin production.}"\\ \\
\noindent
``\texttt{The diagnosis of vascular lesions often requires a combination of clinical examination and sometimes imaging studies.}"\\ \\
\noindent
``\texttt{Dyed lifted polyps can exhibit various morphological features, including lobulated, sessile, or pedunculated appearances.}"\\ \\
\noindent
``\texttt{Endoscopic images of dyed resection margins often show a bright, distinct color outlining the area of resection, contrasting with the surrounding mucosa.}"\\ \\
\noindent
``\texttt{In severe cases, esophagitis may lead to strictures or narrowing of the esophageal lumen, visible during endoscopy.}"\\ \\
\noindent
``\texttt{Endoscopic images of the normal cecum show a well-defined junction with the ascending colon, without any transitional abnormalities.}"\\ \\
\noindent
``\texttt{Endoscopic examination of the normal pylorus shows a lack of any masses, polyps, or other abnormal growths.}"\\ \\
\noindent
``\texttt{The Z line in a normal endoscopy appears intact and well-defined, with no evidence of structural compromise.}"\\ \\
\noindent
``\texttt{Polyps can be classified based on their appearance and histological features, including adenomatous polyps, hyperplastic polyps, or inflammatory polyps.}"\\ \\
\noindent
``\texttt{Ulcerative colitis can be associated with extra-intestinal manifestations, including dermatological, joint, ocular, or hepatobiliary complications.}"\\ \\
}
\subsection{Detailed Base-to-Novel Results}      
Table~\ref{tab:base-to-new} provides detailed base-to-novel evaluation results for each dataset. Overall, MVSL performs on par with state-of-the-art parameter-efficient methods across diverse datasets and outperforms them in most cases.

%%%%%%%%%base-to-new%%%%%%%%%%%%
\begin{table}[h]
\centering
 \caption{\small\textnormal{Accuracy comparison (\%) on Base-to-novel generalization of MVSL with other methods}.}
    \label{tab:base-to-new}
\tablestyle{1pt}{1.1}
\addtolength{\tabcolsep}{1pt}
\resizebox{\columnwidth}{!}{%
\begin{tabular}{lc|c c c c c c c}
\hline
\multirow{1}{*}{Dataset} &  &  {CoOp} & {CoCoOp} &  {KgCoOp} & 
{ProGrad} &{PromptSRC} &
{BiomedCoOp} & {\textbf{MVSL}} \\
\midrule
\multirow{3}{*}{\shortstack[l]{Average on\\  11 datasets}}       
& Base  & 73.85   & 72.26  & 68.36 & 71.67 & 69.01&76.29 & \textbf{80.05} \\
& Novel  & 64.75   & 67.03  & 64.08    & 66.93&70.99 &\textbf{76.29} & 74.58 \\
 & HM & 67.23 & 67.22  & 64.61  & 67.43  &69.99& 76.29 & \textbf{77.22} \\
\midrule
\multirow{3}{*}{BTMRI}     
& Base  & 82.25 & 77.88  & 78.03  & 82.13 &66.16 & 82.42& \textbf{85.69} \\ 
& Novel & 94.51& 94.84  & 95.05  & 94.98 &95.82& \textbf{96.84 }&95.86 \\ 
& HM & 87.95 & 85.53  & 85.69  & 88.09 &78.27& 89.05& \textbf{90.49}  \\
\midrule
\multirow{3}{*}{COVID-QU-Ex}    
& Base   & 75.92 & 77.28  & 75.42 & 75.19 &73.53 & 75.91& \textbf{78.41} \\
& Novel & 90.07 & 87.61   & 89.61  & 90.34   &90.26 & \textbf{91.63} & 90.36  \\
& HM & 82.39 & 82.12  & 81.90  & 82.07&81.04 & 83.03& \textbf{83.96 } \\
\midrule
\multirow{3}{*}{CTKIDNEY}    
& Base    & 82.24 & 81.96  & 81.67  & 83.86&76.63  & \textbf{86.93} &86.15 \\
& Novel & 67.92  & 56.56   & 58.45  & 63.01  &76.77 & 78.94 & \textbf{78.99} \\
& HM  & 74.40 & 66.93  & 68.14  & 71.96  &76.70  & 82.74& \textbf{82.41} \\
\midrule
\multirow{3}{*}{BUSI}    
& Base   & 78.63    & 79.83  & 79.31  & 78.12& 69.23 & 76.92 & \textbf{82.22} \\
& Novel  & 100.00  & 100.00   & 100.00  & 100.00 &100.00  & 100.00 &100.00 \\
& HM & 89.32 &89.92 & 89.66  & 89.06   &81.82 & 88.46& \textbf{90.24} \\
\midrule
\multirow{3}{*}{DermaMNIST}  
& Base & 48.06 & 42.88  & 36.41  & 35.52 &44.90& 54.86 & \textbf{59.39} \\
& Novel & 59.41& 60.66  & 47.31  & 63.28 &51.12& 74.1 & \textbf{75.0} \\
& HM & 53.14 & 50.24  & 41.15  & 45.50  &47.81& 63.04 & \textbf{66.28} \\
\midrule
\multirow{3}{*}{Kvasir}    
& Base & 86.22 & 85.94 & 81.56  & 82.89&85.11 & 86.50& \textbf{87.89} \\
& Novel & 58.06 & 53.95 & 59.00   & 60.45&59.67 & \textbf{61.83 } & 50.11\\
& HM  & 69.39 & 66.29  & 68.47  & 69.91&70.15 & \textbf{72.11} & 63.80 \\
\midrule
\multirow{3}{*}{CHMNIST}       
& Base  & 89.41 & 87.77  & 75.45  & 82.98&80.09 & 88.87& \textbf{93.48} \\
& Novel  & 35.11 & 42.51  & 38.70   & 44.19&47.21 & 42.73& \textbf{45.30} \\
& HM & 50.42  & 57.28  & 51.16  & 57.67  &59.63& 57.71 & \textbf{61.03} \\
\midrule
\multirow{3}{*}{\shortstack[l]{LC25000}}  
& Base  & 90.12 & 88.33  & 88.13  & 90.29 &89.88 & 93.77 &\textbf{96.25}\\
& Novel & 87.55 & 95.02  & 86.44 & 85.47&94.23 & \textbf{97.00} &95.44\\
& HM & 88.82 & 91.55  & 87.28  & 87.81 &92.00& 95.36&\textbf{95.84} \\
\midrule
\multirow{3}{*}{RETINA}        
& Base  & 70.98 & 66.88  & 60.77  & 68.77 &67.98& 68.46 & \textbf{75.99} \\
& Novel  & 56.90 & 65.56  & 54.91  & 58.43 &54.49& 67.72 & \textbf{81.31}\\
& HM & 63.16 & 66.21  & 57.69  & 63.18 &60.49& 68.09 & \textbf{78.56} \\
\midrule
\multirow{3}{*}{KneeXray}           
& Base  & 38.28 & 34.08  & 37.94  & 40.88&37.05 & 44.23 &\textbf{46.98}\\
& Novel & 47.69 & 63.14  & 61.19  & 59.12&61.31 & \textbf{78.35 } & 58.03\\
& HM & 42.47 & 44.27  & 46.84  & 48.34 &46.19& \textbf{56.54} & 51.92 \\
\midrule
\multirow{3}{*}{OCTMNIST}        
& Base & 75.00 & 79.6  & 68.20  & 74.20 &67.73 & 80.33 & \textbf{88.13} \\
& Novel & 50.23 &\textbf{50.47} & 50.13  & 50.02  &50.00& 50.07 & 50.00 \\
& HM & 60.17  & 61.77  & 57.79  & 59.76 &57.53& 61.69 & \textbf{63.80} \\
\bottomrule
\end{tabular}%
}
\end{table}
%%%%%%%%%%%%%%%%%%%%%

\subsection{Detailed Few-shot Results}      
Table~\ref{table:few_all_biomedical} provides detailed few-shot evaluation results for each dataset. Overall, MVSL performs on par with state-of-the-art parameter-efficient methods across diverse datasets and outperforms them in most cases.

%%%%%%%%%%%%%%%%%%%%%%%%%%
\begin{table*}[ht]
\caption{\textbf{Biomedical per-dataset performance} comparison of MVSL with various methods in few-shot setting in terms of classification accuracy (\%).}
\label{table:few_all_biomedical}
\centering
\tablestyle{-5pt}{1.1}
\addtolength{\tabcolsep}{+20pt}
\resizebox{\textwidth}{!}{%
\begin{tabular}{ll|ccccc}
\toprule
\textbf{Dataset} & 
\textbf{Method} & 
$K = 1$ &
$K = 2$ & 
$K = 4$ & 
$K = 8$ &
$K = 16$ \\  \midrule
\multirow{12}{*}{BTMRI}      & BiomedCLIP  & &  & $56.79$ &  &\\ 
& CLIP-Adapter  & $56.80$ & $57.13$ & $56.80$ & $57.15$ & $60.16$ \\
& Tip-Adapter  & $66.66$ & $67.77$ & $76.37$  & $73.75$  & $78.97$ \\
& Tip-Adapter-F  & $59.60$ & $61.94$ & $77.90$ & $79.18$ & $82.27$ \\
& Standard LP  & $62.24$  & $72.45$  & $75.98$  & $77.63$  & $81.24$ \\
& LP++  & $64.72$ & $71.69$ & $75.48$ & $77.11$ & $81.61$ \\
& CoOp  & $63.82$ &  $68.82$ &  $74.68$  &  $79.27$ & $82.37$  \\
& CoCoOp  & $59.47$ &  $64.14$ &  $67.83$  &  $71.69$ & $78.45$  \\
& KgCoOp  & $63.33$ & $70.16$ & $75.40$ & $79.79$ & $81.07$ \\
& ProGrad  & $66.92$ & $71.46$ & $76.24$ & $78.82$ & $82.84$ \\
&  MaPLe  & $38.01$ & $37.02$ & $42.36$ & $50.75$ & $56.22$ \\
& BiomedCoOp  & $65.08$ & $70.57$ & $77.23$ & $78.55$ & $83.30$ \\
\rowcolor{tabhighlight} & MVSL (Ours)  & $66.39$ & $73.97$ & $78.96$ & $80.99$ & $86.70$ \\
\midrule
\multirow{12}{*}{BUSI}      & BiomedCLIP  & &  & $59.75$ &  &\\ 
& CLIP-Adapter  & $61.44$ & $61.01$ & $61.72$ & $61.86$ & $63.55$ \\
& Tip-Adapter  & $62.71$ & $61.44$ & $59.03$  & $55.93$  & $68.78$ \\
& Tip-Adapter-F  & $61.86$ & $56.35$ & $64.54$ & $68.50$ & $71.89$ \\
& Standard LP  & $51.41$  & $47.88$  & $53.38$  & $65.53$  & $68.78$ \\
& LP++  & $51.12$ & $55.50$ & $60.31$ & $66.10$ & $70.05$ \\
& CoOp  & $48.73$ & $53.53$ & $60.17$ & $64.69$ & $69.49$ \\
& CoCoOp  & $52.26$ &  $49.15$ &  $59.75$  &  $65.82$ & $70.2$  \\
& KgCoOp  & $53.39$ & $55.51$ & $62.01$ & $67.37$ & $70.62$ \\
& ProGrad  & $46.33$ & $49.15$ & $62.29$ & $64.83$ & $71.47$ \\
&  MaPLe  & $41.38$ & $33.47$ & $47.74$ & $42.65$ & $45.62$ \\
& BiomedCoOp  & $50.71$ & $50.71$ & $59.32$ & $63.27$ & $70.34$ \\
\rowcolor{tabhighlight} & MVSL (Ours)   & $50.85$ & $51.55$ & $59.32$ & $66.81$ & $73.02$ \\
\midrule
\multirow{12}{*}{COVID-QU-Ex}      & BiomedCLIP  & &  & $43.8$ &  &\\ 
& CLIP-Adapter  & $50.42$ & $43.04$ & $46.28$ & $48.68$ & $49.55$ \\
& Tip-Adapter  & $62.13$ & $58.72$ & $63.84$  & $66.77$  & $73.05$ \\
& Tip-Adapter-F  & $54.89$ & $54.01$ & $69.97$ & $69.89$ & $76.07$ \\
& Standard LP  & $49.91$  & $48.06$  & $60.55$  & $68.29$  & $71.98$ \\
& LP++  & $46.41$ & $56.42$ & $62.32$ & $66.19$ & $72.79$ \\
& CoOp  & $58.82$ & $58.37$  &  $67.03$  & $74.66$  &  $76.37$ \\
& CoCoOp  & $69.36$ &  $68.80$ &  $63.70$  &  $69.36$ & $74.52$  \\
& KgCoOp  & $61.68$ & $54.68$ & $65.91$ & $74.86$ & $75.65$ \\
& ProGrad  & $60.42$ & $64.22$ & $68.56$ & $74.65$ & $74.93$ \\
&  MaPLe  & $35.86$ & $38.99$ & $33.32$ & $36.43$ & $40.89$ \\
& BiomedCoOp  & $72.64$ & $71.53$ & $73.28$ & $76.26$ & $78.72$ \\
\rowcolor{tabhighlight} & MVSL (Ours)  & $71.84$ & $73.46$ & $74.3$ & $76.89$ & $78.96$ \\
\midrule
\multirow{12}{*}{CTKIDNEY}      & BiomedCLIP  & &  & $42.43$ &  &\\ 
& CLIP-Adapter  & $47.17$ & $41.94$ & $42.19$ & $44.64$ & $47.28$ \\
& Tip-Adapter  & $45.85$ & $51.65$ & $55.33$  & $69.89$  & $73.38$ \\
& Tip-Adapter-F  & $46.68$ & $58.99$ & $60.18$ & $75.24$ & $82.07$ \\
& Standard LP  & $43.82$  & $59.35$  & $69.54$  & $78.89$  & $82.50$ \\
& LP++  & $57.70$ & $61.57$ & $65.73$ & $77.06$ & $79.07$ \\
& CoOp  & $54.51$ & $60.57$  &  $68.12$  & $77.40$  &  $83.52$ \\
& CoCoOp  & $47.88$ &  $52.71$ &  $61.07$  &  $73.93$ & $77.70$  \\
& KgCoOp  & $58.92$ & $62.81$ & $68.68$ & $77.43$ & $77.67$ \\
& ProGrad  & $54.65$ & $64.66$ & $67.90$ & $78.23$ & $81.13$ \\
&  MaPLe  & $30.62$ & $38.98$ & $38.00$ & $39.67$ & $51.06$ \\
& BiomedCoOp  & $56.13$ & $64.21$ & $66.50$ & $77.16$ & $83.20$ \\
\rowcolor{tabhighlight} & MVSL (Ours)  & $49.87$ & $61.96$ & $72.21$ & $82.72$ & $88.91$ \\
\bottomrule
\end{tabular}
}
\end{table*}
%%%%%%%%%%%%%%%%%%%%%
% Table Part 2
%%%%%%%%%%%%%%%%%%%%%
\begin{table*}[ht]
\caption*{Table \ref*{table:few_all_biomedical} (continued): \textbf{Biomedical per-dataset performance} comparison of MVSL with various methods in few-shot setting in terms of classification accuracy (\%).}
\centering
\tablestyle{-5pt}{1.1}
\addtolength{\tabcolsep}{+20pt}
\resizebox{\textwidth}{!}{%
\begin{tabular}{ll|ccccc}
\toprule
\textbf{Dataset} & 
 \textbf{Method} & 
$K$ = 1 &
$K$ = 2 & 
$K$ = 4 & 
$K$ = 8 &
$K$ = 16 \\  \midrule
\multirow{12}{*}{Kvasir}      & BiomedCLIP  &   &  & $54.58$ &  &  \\ 
& CLIP-Adapter  & $54.83$ & $54.83$ & $54.83$ & $56.08$ & $56.50$ \\
& Tip-Adapter  & $56.72$ & $60.94$ & $69.61$  & $69.13$  & $74.22$ \\
& Tip-Adapter-F  & $59.19$ & $64.22$ & $69.94$ & $75.86$ & $78.00$ \\
& Standard LP  & $54.30$  & $62.00$  & $72.38$  & $78.88$  & $79.00$ \\
& LP++  & $58.27$ & $60.47$ & $69.36$ & $72.52$ & $75.41$ \\
& CoOp  & $58.2$ &  $64.86$ &  $70.78$  &  $77.14$ & $77.88$  \\
& CoCoOp  & $59.45$ &  $65.50$ &  $68.94$  &  $72.92$ & $75.22$  \\
& KgCoOp  & $61.67$ & $65.67$ & $68.28$ & $72.05$ & $72.95$ \\
& ProGrad  & $60.78$ & $64.70$ & $70.00$ & $76.03$ & $75.88$ \\
&  MaPLe  & $41.06$ & $45.17$ & $53.22$ & $56.03$ & $63.50$ \\
&  BiomedCoOp  & $62.17$ & $67.25$ & $74.08$ & $77.72$ & $78.89$ \\
\rowcolor{tabhighlight} & MVSL (Ours)   & $64.36$ & $69.47$ & $78.19$ & $81.56$ & $83.31$ \\
\midrule
\multirow{12}{*}{CHMNIST}      & BiomedCLIP  & &  & $30.65$ &  &\\ 
& CLIP-Adapter  & $31.27$ & $31.67$ & $33.26$ & $36.48$ & $42.06$ \\
& Tip-Adapter  & $46.14$ & $63.32$ & $70.05$  & $69.57$  & $77.68$ \\
& Tip-Adapter-F  & $52.81$ & $58.90$ & $71.74$ & $74.51$ & $80.43$ \\
& Standard LP  & $58.44$  & $64.42$  & $71.07$  & $76.30$  & $80.34$ \\
& LP++  & $57.18$ & $60.61$ & $67.79$ & $72.40$ & $78.32$ \\
& CoOp  & $57.34$ & $59.68$  &  $68.66$  & $75.00$  &  $79.63$ \\
& CoCoOp  & $49.07$ &  $50.82$ &  $58.58$  &  $66.58$ & $72.16$  \\
& KgCoOp  & $59.02$ & $60.06$ & $68.77$ & $69.50$ & $73.58$ \\
& ProGrad  & $60.15$ & $59.60$ & $69.13$ & $70.99$ & $75.11$ \\
&  MaPLe  & $48.05$ & $57.76$ & $65.87$ & $68.88$ & $71.99$ \\
&  BiomedCoOp  & $59.82$ & $59.79$ & $71.19$ & $74.78$ & $79.05$ \\
\rowcolor{tabhighlight} & MVSL (Ours)  & $63.43$ & $71.12$ & $78.37$ & $84.99$ & $87.43$ \\
\midrule
\multirow{12}{*}{LC25000}      & BiomedCLIP  & &  & $50.03$ &  &\\ 
& CLIP-Adapter  & $54.83$ & $53.47$ & $52.91$ & $56.33$ & $57.56$ \\
& Tip-Adapter  & $75.37$ & $72.73$ & $83.32$  & $87.25$  & $89.17$ \\
& Tip-Adapter-F  & $74.21$ & $71.82$ & $79.57$ & $90.41$ & $92.35$ \\
& Standard LP  & $74.50$  & $78.40$  & $85.30$  & $90.24$  & $92.77$ \\
& LP++  & $63.05$ & $71.42$ & $82.61$ & $89.14$ & $92.58$ \\
& CoOp  & $71.90$ & $76.55$  &  $84.66$  & $87.50$  &  $92.19$ \\
& CoCoOp  & $63.66$ &  $71.76$ &  $77.44$  &  $85.57$ & $87.38$  \\
& KgCoOp  & $71.80$ & $75.18$ & $82.10$ & $84.63$ & $86.79$ \\
& ProGrad  & $72.48$ & $74.76$ & $84.72$ & $87.86$ & $90.70$ \\
&  MaPLe  & $67.13$ & $69.80$ & $76.73$ & $82.19$ & $86.73$ \\
&  BiomedCoOp  & $77.56$ & $77.74$ & $85.60$ & $88.77$ & $92.68$ \\
\rowcolor{tabhighlight} & MVSL (Ours)  & $81.22$ & $78.17$ & $89.53$ & $92.32$ & $95.32$ \\
\midrule
\multirow{12}{*}{RETINA}      & BiomedCLIP  & &  & $26.26$ &  &\\ 
& CLIP-Adapter  & $25.49$ & $25.49$ & $26.07$ & $25.84$ & $26.05$ \\
& Tip-Adapter  & $26.52$ & $31.07$ & $43.42$  & $48.08$  & $54.23$ \\
& Tip-Adapter-F  & $39.53$ & $33.07$ & $47.37$ & $56.07$ & $62.85$ \\
& Standard LP  & $39.35$  & $46.03$  & $51.31$  & $53.94$  & $62.27$ \\
& LP++  & $35.77$ & $39.37$ & $46.95$ & $53.44$ & $60.62$ \\
& CoOp  & $35.02$ & $35.26$  &  $42.22$  & $51.87$  &  $59.38$ \\
& CoCoOp  & $32.94$ &  $36.43$ &  $39.75$  &  $48.45$ & $53.91$  \\
& KgCoOp  & $33.54$ & $35.17$ & $42.61$ & $49.97$ & $51.18$ \\
& ProGrad  & $33.49$ & $36.49$ & $43.09$ & $52.26$ & $50.47$ \\
&  MaPLe  & $27.73$ & $31.07$ & $30.89$ & $41.80$ & $53.78$ \\
&  BiomedCoOp  & $36.64$ & $38.67$ & $45.58$ & $56.47$ & $61.28$ \\
\rowcolor{tabhighlight} & MVSL (Ours)  & $40.72$ & $42.90$ & $54.05$ & $61.04$ & $68.11$ \\
\bottomrule
\end{tabular}
}
\end{table*}
%%%%%%%%%%%%%%%%%%%%%

% Table Part 3
%%%%%%%%%%%%%%%%%%%%%
\begin{table*}[ht]
\caption*{Table \ref*{table:few_all_biomedical} (continued): \textbf{Biomedical per-dataset performance} comparison of MVSL with various methods in few-shot setting in terms of classification accuracy (\%).}
\centering
\tablestyle{-5pt}{1.1}
\addtolength{\tabcolsep}{+20pt}
\resizebox{\textwidth}{!}{%
\begin{tabular}{ll|ccccc}
\toprule
\textbf{Dataset} & 
 \textbf{Method} & 
$K$ = 1 &
$K$ = 2 & 
$K$ = 4 & 
$K$ = 8 &
$K$ = 16 \\  \midrule
\multirow{12}{*}{KneeXray}      & BiomedCLIP  & &  & $29.53$ &  &\\ 
& CLIP-Adapter  & $29.00$ & $28.66$ & $28.96$ & $28.80$ & $29.08$ \\
& Tip-Adapter  & $29.04$ & $33.55$ & $24.19$  & $25.76$  & $33.17$ \\
& Tip-Adapter-F  & $30.01$ & $28.38$ & $26.59$ & $26.46$ & $27.67$ \\
& Standard LP  & $26.02$  & $26.57$  & $27.83$  & $22.20$  & $23.97$ \\
& LP++  & $21.25$ & $26.40$ & $28.92$ & $23.75$ & $26.38$ \\
& CoOp  & $24.96$ & $25.89$  &  $23.85$  & $26.23$  &  $28.48$ \\
& CoCoOp  & $25.42$ &  $28.85$ &  $30.66$  &  $21.78$ & $24.86$  \\
& KgCoOp  & $29.07$ & $28.14$ & $22.44$ & $23.37$ & $24.80$ \\
& ProGrad  & $30.09$ & $23.83$ & $23.95$ & $24.78$ & $26.27$ \\
&  MaPLe  & $23.41$ & $22.67$ & $22.56$ & $24.78$ & $25.87$ \\
&  BiomedCoOp  & $36.13$ & $37.72$ & $35.91$ & $37.7$ & $39.69$ \\
\rowcolor{tabhighlight} & MVSL (Ours) & $37.54$ & $40.56$ & $39.84$ & $42.65$ & $43.74$ \\
\midrule
\multirow{12}{*}{OCTMNIST}      & BiomedCLIP  & &  & $30.00$ &  &\\ 
& CLIP-Adapter  & $44.00$ & $49.73$ & $49.96$ & $49.50$ & $52.73$ \\
& Tip-Adapter  & $32.36$ & $33.8$ & $38.10$  & $53.93$  & $53.33$ \\
& Tip-Adapter-F  & $46.66$ & $53.93$ & $55.20$ & $65.00$ & $72.50$ \\
& Standard LP  & $47.25$  & $54.21$  & $61.00$  & $65.85$  & $69.40$ \\
& LP++  & $47.24$ & $53.18$ & $59.02$ & $63.69$ & $68.35$ \\
& CoOp  & $52.63$ & $53.57$  &  $53.37$  & $63.67$  &  $65.47$ \\
& CoCoOp  & $49.33$ &  $50.93$ &  $48.57$  &  $55.40$ & $60.67$  \\
& KgCoOp  & $50.63$ & $50.53$ & $52.97$ & $61.03$ & $62.80$ \\
& ProGrad  & $51.40$ & $55.33$ & $55.07$ & $62.17$ & $63.33$ \\
& MaPLe  & $26.63$ & $34.00$ & $30.17$ & $30.53$ & $33.63$ \\
& BiomedCoOp  & $51.83$ & $55.03$ & $54.73$ & $58.87$ & $66.93$ \\
\rowcolor{tabhighlight} & MVSL (Ours)  & $50.77$ & $56.20$ & $57.67$ & $67.23$ & $74.00$ \\
\midrule
\multirow{12}{*}{\textbf{Average}}      & BiomedCLIP  & &  & $42.38$ &  &\\ 
& CLIP-Adapter  & $45.53$ & $44.70$  & $45.30$  & $46.54$  & $48.46$  \\
& Tip-Adapter & $50.35$ & $53.50$ & $58.33$  & $62.01$   & $67.60$ \\
& Tip-Adapter-F & $52.55$ & $54.17$ & $62.30$  & $68.12$ & $68.12$ \\
& Standard LP & $48.91$  & $55.82$  & $62.12$ & $67.33$  & $70.81$\\
& LP++ & $49.27$ & $55.88$  & $61.30$ & $65.48$ & $70.09$ \\
& CoOp  & $52.59$ & $55.71$  &  $61.35$  & $67.74$ &  $71.48$ \\
& CoCoOp  & $50.88$  &  $53.91$  &  $57.63$ &  $63.15$  & $67.51$   \\
& KgCoOp & $54.31$  & $55.79$  & $60.92$  & $66.00$ & $67.71$ \\
& ProGrad  & $53.67$  & $56.42$  & $62.10$  & $67.06$  & $69.21$  \\
& MaPLe  & $37.99$ & $40.89$ &  $44.09$ & $47.37$ & $52.93$ \\
& BiomedCoOp  & $56.87$ & $59.32$ & $64.34$ & $68.96$  & $73.41$ \\
\rowcolor{tabhighlight} & MVSL (Ours)  & $58.41$ & $62.17$ & $68.08$ & $73.10$ & $77.13$ \\
\bottomrule
\end{tabular}
}
\end{table*}
 
\end{document}